\documentclass{article} 
\usepackage{iclr2025_conference,times}

\usepackage{amsmath,amsfonts,bm}

\def\eqref#1{equation~\ref{#1}}

\def\1{\bm{1}}

\DeclareMathAlphabet{\mathsfit}{\encodingdefault}{\sfdefault}{m}{sl}
\SetMathAlphabet{\mathsfit}{bold}{\encodingdefault}{\sfdefault}{bx}{n}

\usepackage{hyperref}
\usepackage{url}

\usepackage{enumitem}            
\usepackage{booktabs}       
\usepackage{microtype}      
\usepackage{bigstrut,multirow,rotating}
\usepackage{wrapfig}
\usepackage{subcaption}
\usepackage{adjustbox}
\newtheorem{definition}{Definition}
\newtheorem{observation}{Observation}

\usepackage{array}

\title{Dataset Ownership Verification in Contrastive Pre-trained Models}

\author{Yuechen Xie\textsuperscript{\rm 1}, Jie Song\textsuperscript{\rm 1}\thanks{Corresponding authors.},  Mengqi Xue\textsuperscript{\rm 2}, Haofei Zhang\textsuperscript{\rm 1} \\
\bf Xingen Wang\textsuperscript{\rm 1,4}, Bingde Hu\textsuperscript{\rm 1,4}, Genlang Chen\textsuperscript{\rm 3}, Mingli Song\textsuperscript{\rm 1}\\
 $^1$Zhejiang University, 
 $^2$Hangzhou City University\\
 $^3$NingboTech University, 
 $^4$Bangsheng Technology Co., Ltd.\\
 \small\{xyuechen, sjie, haofeizhang, newroot,  tonyhu, cgl, brooksong\}@zju.edu.cn, mqxue@hzcu.edu.cn\\
 }

\iclrfinalcopy 
\begin{document}

\maketitle


\begin{abstract}
High-quality open-source datasets, which necessitate substantial efforts for curation, has become the primary catalyst for the swift progress of deep learning. Concurrently, protecting these datasets is paramount for the well-being of the data owner. Dataset ownership verification emerges as a crucial method in this domain, but existing approaches are often limited to supervised models and cannot be directly extended to increasingly popular unsupervised pre-trained models.
In this work, we propose the first dataset ownership verification method tailored specifically for self-supervised pre-trained models by contrastive learning. Its primary objective is to ascertain whether a suspicious black-box backbone has been pre-trained on a specific unlabeled dataset, aiding dataset owners in upholding their rights. 
The proposed approach is motivated by our empirical insights that when models are trained with the target dataset, the unary and binary instance relationships within the embedding space exhibit significant variations compared to models trained without the target dataset. We validate the efficacy of this approach across multiple contrastive pre-trained models including SimCLR, BYOL, SimSiam, MOCO v3, and DINO. The results demonstrate that our method rejects the null hypothesis with a $p$-value markedly below $0.05$, surpassing all previous methodologies. Our code is available at \url{https://github.com/xieyc99/DOV4CL}.
\end{abstract}

\section{Introduction}

The success of deep learning is greatly dependent on the the availability of high-quality open-source datasets, which empower researchers and developers to train and test their models and algorithms. Presently, the majority of public datasets~\cite{deng2009imagenet,krizhevsky2009learning,netzer2011reading} are designated exclusively for academic purposes, with commercial use prohibited without explicit permission. Therefore, preventing the stealing of public datasets holds significant importance for the benefit of the data owners.

Numerous traditional techniques exist for data security, including encryption~\cite{boneh2001identity,khamitkarsurvey}, differential privacy ~\cite{dwork2006differential,abadi2016deep}, and digital watermarking~\cite{cox2002digital,podilchuk2001digital,kadian2021robust}. However, these methods fall short in protecting the copyrights of open-source datasets, as they either impede dataset accessibility or necessitate the knowledge of the training process of potentially suspicious models. Recently,
dataset ownership verification~(DOV)~\cite{guo2023domain,li2022untargeted,li2023black} emerges as a novel defense measure to deter dataset theft. It allows defenders, \textit{i.e.,} dataset owners, to demonstrate whether suspects have infringed upon their rights by ascertaining whether a suspicious black-box backbone has been pre-trained on their datasets. 
However, as most existing DOV techniques are designed solely for supervised models where verification relies on distances between data points and decision boundaries~~\cite{li2018decision,karimi2019characterizing,karimi2020decision}, they are not directly applicable to recently increasing popular self-supervised pre-trained models~~\cite{chen2020simple,chen2021exploring,chen2021empirical} due to the absence of the well-defined decision boundaries.

\begin{figure}[t]
    \centering
    \includegraphics[width=\textwidth, trim=0.3cm 0.3cm 0.3cm 0.3cm, clip]{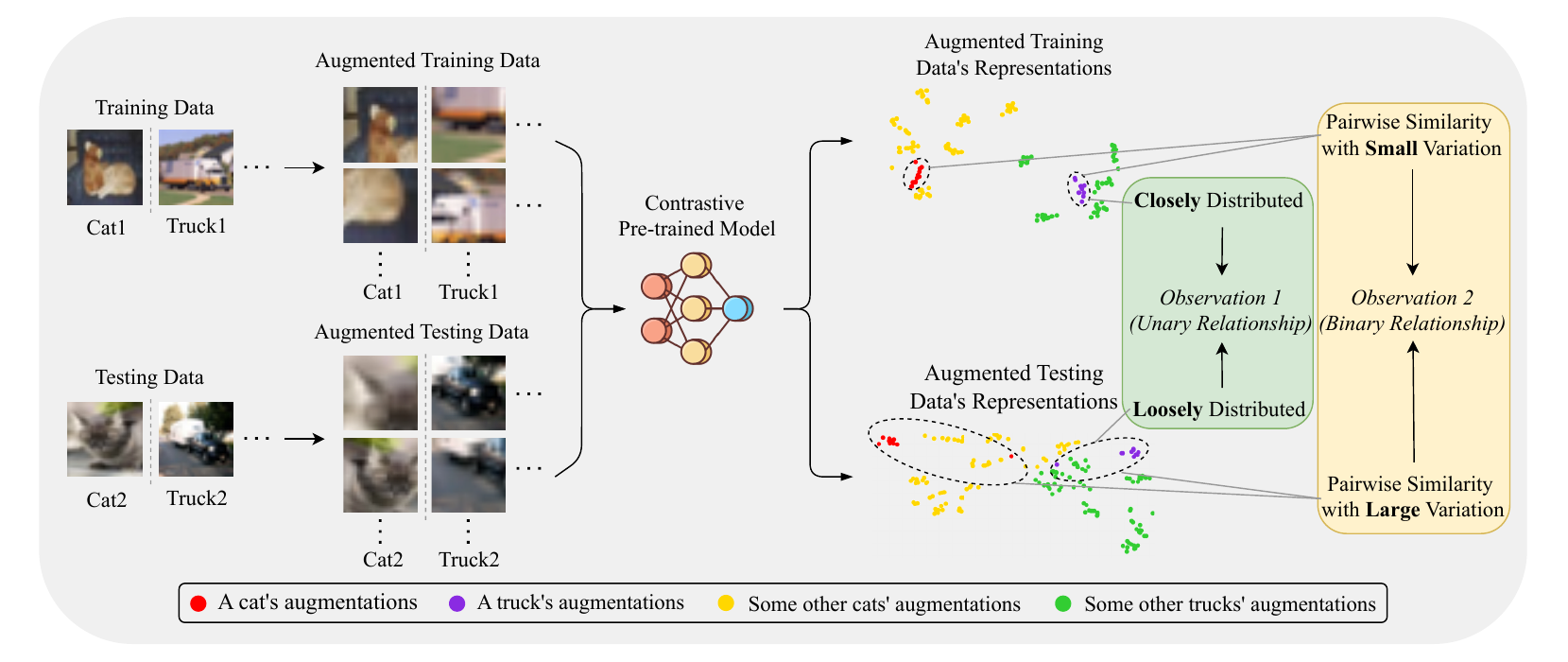}
    \caption{The overview of the two key observations. The representations are visualized using t-SNE. The encoder is a ResNet18 pre-trained on CIFAR10 with BYOL~\cite{grill2020bootstrap}. 
    }
    \vspace{-10pt}
    \label{fig:motivations}
\end{figure}

In this work, we present, to the best of our knowledge, the first DOV method for contrastive pre-trained models. It aids defenders in validating whether suspicious models have been illicitly pre-trained on their public datasets. Given a third-party suspicious model that might be pre-trained on the protected dataset without authorization, we focus on the black-box setting where defenders have no information about other training configurations (\textit{e.g.}, loss function and model architecture) of the model and can only access model via Encoder as a Service (EaaS)~\cite{sha2023can,liu2022stolenencoder}. It means defenders can only retrieve feature vectors via model API.
The proposed approach is formulated upon two key observations, as shown in Figure \ref{fig:motivations}. (1)  \textit{Unary relationship}: encoders pre-trained through contrastive learning generate remarkably more similar representations for augmentations of the same seen samples at the training phase than the unseen samples. (2) \textit{Binary relationship}: the pairwise similarity between the seen samples doesn't significant change after data augmentations.

We define the differences in unary and binary relationships between seen and unseen samples as the \textit{contrastive relationship gap} of the suspicious model. Defenders can endeavor to activate this gap in the suspicious encoder by employing their own public datasets, in order to ascertain whether the suspect's encoder was pre-trained on their data. 
More specifically, as illustrated in Figure \ref{fig:flowChart}, the proposed DOV technique comprises three steps: (1) pre-training a \textit{shadow} encoder devoid of the public dataset of the defender; (2) utilizing multi-scale augmentation to compute the contrastive relationship gaps of the suspect encoder and the shadow encoder; (3) conducting hypothesis testing on the contrastive relationship gaps of the two encoders to determine whether the suspect encoder has been pre-trained on the defender's public dataset.

In summary, the principal contributions of this paper are threefold: (1) we discern that when models are trained with the target dataset, the unary and binary instance relationships within the embedding space demonstrate noteworthy disparities in comparison to models trained without the target dataset; (2) we introduce the concept of the contrastive relationship gap, which, to the best of our knowledge, represents the first DOV technique for contrastive pre-trained models; (3) comprehensive experiments showcase that our approach refutes the null hypothesis with a $p$-value significantly below $0.05$, surpassing all preceding studies.

\section{Related Work}
\label{gen_inst}

\paragraph{Data Protection.}
Dataset ownership verification is an emerging field in data security. Typically, it involves embedding watermarks into the original dataset~\citep{guo2023domain,li2022untargeted,li2023black,tang2023did}. Models trained on the watermarked dataset will incorporate a pre-designed backdoor, allowing defenders to verify data ownership simply by triggering the model’s backdoor. However, current DOV methods primarily target supervised models and require altering the original dataset's distribution to inject watermarks, which makes it susceptible to various watermark removal mechanisms~\citep{chen2021refit,liu2021wdnet,sun2023defending,kwon2021defending,hayase2021spectre}. 
The proposed method demonstrates that, for contrastive learning, dataset ownership can be efficiently verified without modifying the original dataset. 

Dataset inference~\cite{maini2021dataset} is a state-of-the-art defense against model stealing~\citep{sha2023can,sanyal2022towards,shen2022model}. It does not require retraining the model or embedding watermarks within the dataset, which reduces the time cost significantly while preserving the original distribution of the data. The latest dataset inference method~\cite{dziedzic2022dataset} has expanded its application to self-supervised learning. Although it's primarily aimed at encoder theft, it can also be directly used for dataset ownership verification. However, it necessitates inferring the entire training set to model the output features of all data from both the training and testing sets. It is prohibitively time-consuming for large datasets, such as ImageNet~\citep{deng2009imagenet}. In contrast, our method achieves accurate verification using only a small fraction of the dataset. For instance, on ImageNet, we use only 0.1\% of the training set for verification.

Membership inference~\cite{shokri2017membership,choquette2021label,carlini2022membership,hu2022membership} aims to determine whether an input was part of the model's training dataset. EncoderMI~\cite{liu2021encodermi} is a powerful method specifically designed for membership inference on encoders pre-trained via contrastive learning, which takes advantage of the overfitting tendencies of the image encoder. However, it directly trains the inferencer on high-dimensional representations that contain a large amount of redundant information, which leads to a heavy computational cost and increased training difficulty. In contrast, our method extracts the most critical information for verification from the representations, namely contrastive relationship gap, achieving effective verification without the need to train an inferencer.

Inspired by Proof of Learning (PoL)~\cite{jia2021proof,fang2023proof,zhao2024proof}, Proof of Training Data (PoTD)~\cite{choi2024tools} is proposed to assist third-party auditor in validating which data were used to train models. It helps develop practical and robust tools for accountability in the large-scale development of artificial intelligence models. However, it entails substantial verification costs, as the model trainer (suspect) is required to disclose detailed training records to the verifier, including training data, training code, and intermediate checkpoints. In practical scenarios, if the models trained by the suspect possess significant commercial value, the suspect is seldom willing to comply with such disclosures. Our setup is more reflective of real-world scenarios, where the model is a black box, and the defender can only access its API. 


\paragraph{Contrastive Learning.}

Contrastive learning~\cite{chen2020simple,chen2021exploring,chen2021empirical,caron2020unsupervised,albelwi2022survey,he2020momentum} aims to pre-train image encoders on unlabeled data by leveraging the supervisory signals inherent in the data itself, with these pre-trained encoders being applicable to numerous downstream tasks. The central idea of contrastive learning is to enable the encoder to produce similar feature vectors for a pair of augmentations derived from the same input image (positive samples), and distinct feature vectors for augmentations derived from different input images (negative samples). Classical approaches like SimCLR~\cite{chen2020simple}, MoCo~\cite{he2020momentum}, SwAV~\cite{caron2020unsupervised}, utilize both positive samples (for feature alignment) and negative samples (for feature uniformity). Surprisingly, researchers notice that contrastive learning can also work well by only aligning positive samples, such as BYOL~\cite{grill2020bootstrap} and DINO~\cite{caron2021emerging}. We follow some literatures~\cite{albelwi2022survey,gao2022disco} to coin these methods as a special type of contrastive learning, or contrastive learning without negatives. We make no strict distinction between these concepts here due to the clear context in this work. Our method is designed to protect the unlabeled datasets used in contrastive learning, thereby securing and fostering healthy development in this field.

\section{The Proposed Method}
\label{headings}

\subsection{Problem Formulation}
\label{problem_formulate}

In this study, we focus on the dataset ownership verification task in black-box scenarios. The problem involves two key player: the \textit{defender} and the \textit{suspect}. The defender, assuming the role of the dataset provider, endeavors to ascertain whether the suspect model, $\mathcal {M}_{sus}$, has been unlawfully trained on his public dataset $\mathcal D_{pub}$. $\mathcal {M}_{sus}$ can be classified into four scenarios based on its training datasets: 
\normalsize{\textcircled{\scriptsize{\textbf{1}}}}\normalsize\;$\mathcal {M}_{sus}$ is exclusively trained on the  public dataset $\mathcal {D}_{pub}$ of the defender, indicating the occurrence of dataset misappropriation; 
\normalsize{\textcircled{\scriptsize{\textbf{2}}}}\normalsize\;$\mathcal {M}_{sus}$ is trained on a dataset  that encompasses the designated public dataset $\mathcal {D}_{pub}$ along with an additional dataset $\mathcal {D}_{alt}$, signifying dataset misappropriation, albeit posing a more challenging DOV task than case \normalsize{\textcircled{\scriptsize{\textbf{1}}}}\normalsize\;due to the presence of $\mathcal {D}_{alt}$; 
\normalsize{\textcircled{\scriptsize{\textbf{3}}}}\normalsize\;$\mathcal {M}_{sus}$ is trained on an unrelated dataset $\mathcal {D}_{unre}$ outside the scope of the defender's public dataset, indicating the innocence of the suspect;  
\normalsize{\textcircled{\scriptsize{\textbf{4}}}}\normalsize\;$\mathcal {M}_{sus}$ is trained on an alternative dataset $\mathcal {D}_{alt}$ that bears significant resemblance yet doesn't overlap with the public dataset $\mathcal {D}_{pub}$, suggesting the innocence of the suspect, albeit posing a more arduous DOV challenge than case \normalsize{\textcircled{\scriptsize{\textbf{3}}}}\normalsize. 
These four scenarios encompass nearly every conceivable real-world circumstance.

\subsection{Contrastive Relationship Gap}
\label{our_method}

\subsubsection{Observations and Definitions}

In contrastive learning, a pivotal training objective for encoders is to maximize the similarity between the representations of positive samples, which are different augmentations of the same training image. This training approach leverages the neural network's memory capacity, prompting the encoder to retain the features of the training data. As a result, we derive the following two significant insights:

\begin{observation}[\textbf{Unary Relationship}]
Contrastive pre-trained encoders can produce more alike representations for the same seen samples' augmentations during pre-training than unseen samples.
\label{ob1}
\end{observation}

\begin{observation}[\textbf{Binary Relationship}]
The pairwise similarity between the seen samples’ representations hardly change after augmentations, unlike with unseen samples during pre-training.
\label{ob2}
\end{observation}

\vspace{-10pt}

We characterize the disparity between familiar and unfamiliar data encountered during the training phase as the encoder's \textit{contrastive relationship gap}, a metric that can aid defenders in discerning whether the queried encoder has been pre-trained on their dataset. The precise definition is as follows:

\begin{definition}[\textbf{Contrastive Relationship Gap}]
Given a contrastive pre-trained encoder $\mathcal{M}$ and a dataset $\mathcal{D}$, the contrastive relationship gap of $\mathcal{M}$ is defined as: 
\begin{equation}
d\big( {\mathcal{D},\hat{\mathcal{D}},\mathcal{M},T} \big) = \bigg\{ s_i - \hat{s}_i \Big| i \in \big[1,|\mathcal{S}| \big], s_i \in \mathcal{S}\big( {\mathcal{D},\mathcal{M},T} \big), \hat{s}_i \in \mathcal{S}\big( {\hat{\mathcal{D}},\mathcal{M},T} \big) \bigg\}
\label{eq_CMD} 
\end{equation}
where $\hat{\mathcal{D}}$ is a dataset that $\mathcal{M}$ has not been pre-trained on. \(T( \cdot )\) denotes an augmentation function. \(\mathcal{S}\left({\cdot,\cdot,\cdot}\right)\) is a similarity set. \(|\mathcal{S}|\) is the total number of samples in \(\mathcal{S} ( {\mathcal{D},\mathcal{M},T} ) \).
\end{definition}
A larger mean of contrastive relationship gap suggests that $\mathcal{M}$ is more likely to have been pre-trained on $\mathcal{D}$. According to Observation \ref{ob1} and Observation \ref{ob2},  \(\mathcal{S}\) consists of unary relationship similarity set \(\mathcal{S}_U\) and binary relationship similarity set \(\mathcal{S}_B\).

\subsubsection{The Calculation of \texorpdfstring{$\mathcal{S}_U$}{Lg} and \texorpdfstring{$\mathcal{S}_B$}{Lg}}


Random cropping is a commonly used data augmentation technique in contrastive learning~\cite{chen2020simple,chen2021exploring,chen2021empirical}, which can enhance the model's generalization ability significantly. In this paper, we use multi-scale random cropping to capture both global and local features of objects. Specifically, we design the \(T\) in Eq.(\ref{eq_CMD}) as a multi-scale augmentation function \({T^{ms}} = \{{T^g},{T^l}\}\), hoping to activate the encoder's contrastive relationship gap from various dimensions. \({{T^g}}\) is the global augmentation function responsible for larger regions, while \(T^l\) is the local augmentation function focusing on smaller regions. Through \( T^{ms} \), we calculate \(\mathcal{S}_U\) and \(\mathcal{S}_B\) at multi-scale. Their definitions are as follows:

\begin{definition}[\textbf{Unary Relationship Similarity Set}]
Given an encoder $\mathcal{M}$ and a dataset $\mathcal{D}$, the unary relationship similarity set is defined as:
\begin{equation}
{\mathcal{S}_U}\big(\mathcal{D},\mathcal{M},T^{ms}\big) = \{S_U^{gg},S_U^{ll},S_U^{gl}\}
\label{eq_IMAS}
\end{equation}
where \({S_U^{gg}}\), \({S_U^{ll}}\), and \({S_U^{gl}}\) respectively denote the unary relationship similarity between global and global views, local and local views, and global and local views. 
\end{definition}
The specific formulas are as follows:
\begin{equation}
S_U^{gg} = \frac{2}{{|\mathcal{D}|M(M - 1)}}\sum\nolimits_{i = 1}^{|\mathcal{D}|} {\sum\nolimits_{m = 1}^M {\sum\nolimits_{n = m + 1}^M {sim\Big( {\mathcal{M} \big({ T^g_m{\left( {{x_i}} \right)}}\big),\mathcal{M} \big( T^g_n{{\left( {{x_i}} \right)}}\big)}\Big) } }}
\end{equation}
\begin{equation}
S_U^{ll} = \frac{2}{{|\mathcal{D}|N(N - 1)}}\sum\nolimits_{i = 1}^{|\mathcal{D}|} {\sum\nolimits_{m = 1}^N {\sum\nolimits_{n = m + 1}^N {sim\Big( {\mathcal{M} \big({ T^l_m{\left( {{x_i}} \right)}}\big),\mathcal{M} \big( T^l_n{{\left( {{x_i}} \right)}}\big)}\Big)} } }
\end{equation}
\begin{equation}
S_U^{gl} = \frac{1}{{|\mathcal{D}|MN}}\sum\nolimits_{i = 1}^{|\mathcal{D}|} {\sum\nolimits_{m = 1}^M {\sum\nolimits_{n = 1}^N {sim\Big( {\mathcal{M} \big({ T^g_m{\left( {{x_i}} \right)}}\big),\mathcal{M} \big( T^l_n{{\left( {{x_i}} \right)}}\big)}\Big)} } }
\end{equation}

where \({{x_i} \in \mathcal{D}}\), and \(|\mathcal{D}|\) is the total number of samples in dataset \(\mathcal{D}\). \(M\) and \(N\) are the execution number for \({{T^g}}\) and \({{T^l}}\), respectively. \({{T^g_m}{{\left({x_i}\right)}}}\) denotes the \(m\)-th augmentation of \({{x_i}}\) by \({{T^g}}\), similarly for \({{T^g_n}{{\left({x_i}\right)}}},{{T^l_m}{{\left({x_i}\right)}}},{{T^l_n}{{\left({x_i}\right)}}}\). \(sim\left({\cdot,\cdot}\right)\) represents the cosine similarity function.

\begin{definition}[\textbf{Binary Relationship Similarity Set}]
Similar to \(\mathcal{S}_U\), given an encoder $\mathcal{M}$ and a dataset $\mathcal{D}$, the binary relationship similarity set is defined as:
\begin{equation}
{\mathcal{S}_B}\big(\mathcal{D},\mathcal{M},T^{ms}\big) = \{S_B^{gg},S_B^{ll},S_B^{gl}\} 
\label{eq_CMAGS}
\end{equation}
where \({S_B^{gg}}\), \({S_B^{ll}}\), and \({S_B^{gl}}\) is the binary relationship similarity between global and global views, local and local views, and global and local views respectively. 
\end{definition}
We first introduce the binary relationship set \(\mathcal{G}\). It includes the pairwise similarity between the augmented images' representations, denoted as:
\begin{equation}
{\mathcal{G}( \mathcal{D},\mathcal{M},T )} = \bigg\{ {sim\Big( {\mathcal{M} \big({ T{\left( {{x_i}} \right)}}\big),\mathcal{M} \big( T{{\left( {{x_j}} \right)}}\big)}\Big)} \Big| i \in \big[ {1,|\mathcal{D}|} \big],j \in \big( {i,|\mathcal{D}|} \big]\Big.\bigg\}
\label{eq_CAG}
\end{equation}
where \(sim\left({\cdot,\cdot}\right)\) denotes the cosine similarity function, with \({{x_i},{x_j} \in \mathcal{D}}\), \(|\mathcal{D}|\) is the total number of samples in dataset \(\mathcal{D}\), and \(T( \cdot )\) represents the augmentation function. By substituting the augmentation functions \({{T^g}}\) and \({{T^l}}\) into Eq.(\ref{eq_CAG}), we obtain the binary relationship set \({{\mathcal{G}^g}}\) and \({{\mathcal{G}^l}}\) at respective scales. Below, we formally present the specific formulas for \({S_B^{gg}}\), \({S_B^{ll}}\), and \({S_B^{gl}}\):
\begin{equation}
S_B^{gg} =  - \frac{2}{{M( {M - 1} )}}\sum\nolimits_{m = 1}^M {\sum\nolimits_{n = m + 1}^M {f( {\mathcal{G}_m^g,\mathcal{G}_n^g} )} }
\end{equation}
\begin{equation}
S_B^{ll} =  - \frac{2}{{N( {N - 1} )}}\sum\nolimits_{m = 1}^N {\sum\nolimits_{n = m + 1}^N {f( {\mathcal{G}_m^l,\mathcal{G}_n^l} )} }
\end{equation}
\begin{equation}
S_B^{gl} =  - \frac{1}{{MN}}\sum\nolimits_{m = 1}^M {\sum\nolimits_{n = 1}^N {f( {\mathcal{G}_m^g,\mathcal{G}_n^l} )} }
\end{equation}
where \(f\left({\cdot,\cdot}\right)\) is a distance measurement function, which is implemented as the mean absolute error in this paper. \(M\) and \(N\) are the execution number for \({{T^g}}\) and \({{T^l}}\), respectively. \({\mathcal{G}_m^g}\) represents the \(m\)-th binary relationship set based on \({{T^g}}\), similarly for \({\mathcal{G}_n^{g}}\), \({\mathcal{G}_m^{l}}\) and \({\mathcal{G}_n^{l}}\).

Using unary relationship similarity set \(\mathcal{S}_U\) and binary relationship similarity set \(\mathcal{S}_B\), we can determine the contrastive relationship gap \(d\) of the encoder $\mathcal{M}$ as follows:
\begin{equation}
d = \left\{ \sum\nolimits_* {(S_U^{*}-\hat{S}_U^{*})\cdot I(S_U^{*}>\hat{S}_U^{*}) }, \sum\nolimits_* {(S_B^{*}-\hat{S}_B^{*})}\cdot I(S_B^{*}>\hat{S}_B^{*}) \right\}
\label{eq_CMD_2}
\end{equation}
where \(*\in\{gg,ll,gl\}\), 
\({S}\) and \(\hat{S}\) come from \(\mathcal{S} ( {\mathcal{D},\mathcal{M},T} ) \) and \(\mathcal{S} ( {\hat{\mathcal{D}},\mathcal{M},T} ) \) in Eq.(\ref{eq_CMD}), respectively.
\( I(\cdot) \) is the function returning \(a\) if the input statement is true or returning 1 if the input statement is false. \(a\) is a hyperparameter with a default value of 1. As \(a\) increases, the contrastive relationship gap of encoder \(\mathcal{M}\) between \(\mathcal{D}\) and \(\hat{\mathcal{D}}\) becomes larger.

\subsubsection{The Complete Process}

\begin{figure}[t]
\centering
\includegraphics[width=1.0\textwidth, trim=0.7cm 0.1cm 0.7cm 0.1cm, clip]{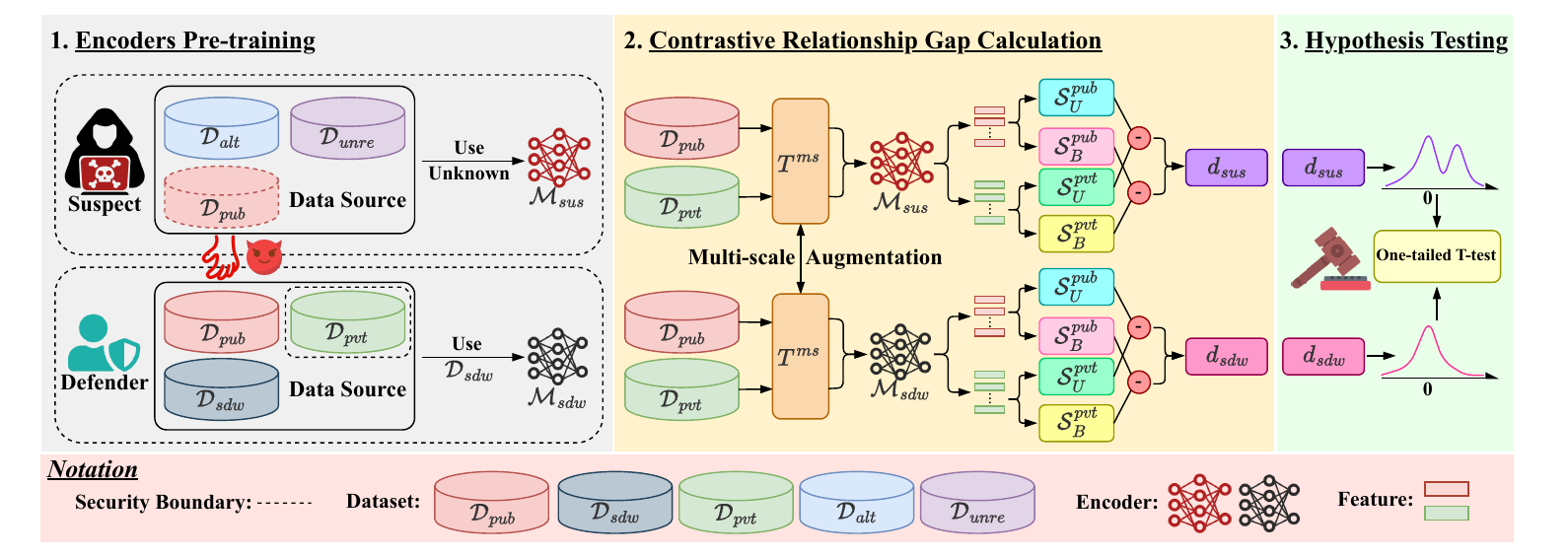} 
\caption{The overview of our method (best viewed under color conditions).}
\vspace{-15pt}
\label{fig:flowChart}
\end{figure}

We propose a method of  dataset ownership verification by contrastive relationship gap.
Figure \ref{fig:flowChart} displays the entire process of our method, divided into three stages: 

(1) pre-training a shadow encoder \(\mathcal{M}_{sdw}\) on a shadow dataset \(\mathcal{D}_{sdw}\) to compare with \(\mathcal{M}_{sus}\); 

(2) performing \(K\) samplings on \( \mathcal{D}_{pub} \) and \( \mathcal{D}_{pvt} \) (a defender's private dataset which isn't publicly available, and \(\mathcal{M}_{sus}\) has not been trained on it), that represent \( \mathcal{D} \) and \( \hat{\mathcal{D}} \) in Eq.(\ref{eq_CMD}), respectively. The sampling sizes are \( k_{pub} \) and \( k_{pvt} \) respectively, resulting in the subsets \(\{ \mathcal{D}_{pub}^1, \cdots ,\mathcal{D}_{pub}^K\} \) and \(\{ \mathcal{D}_{pvt}^1, \cdots ,\mathcal{D}_{pvt}^K\} \). Then using these subsets calculate the contrastive relationship gaps \({d}_{sus} =  {d}_{sus}^1 \cup \cdots \cup {d}_{sus}^K \) and \({d}_{sdw} =  {d}_{sdw}^1 \cup \cdots \cup {d}_{sdw}^K \) of \(\mathcal{M}_{sus}\) and \(\mathcal{M}_{sdw}\), respectively; 

(3) One-tailed pair-wise T-test~\cite{hogg2013introduction} is conducted on \({d}_{sus}\) and \({d}_{sdw}\). The null hypothesis, \({H_0}\), posits that the mean of \({d}_{sus}\) is less than or equal to that of \({d}_{sdw}\), while the alternative hypothesis, denoted as \({H_1}\), posits that the mean of \({d}_{sus}\) is greater than the mean of \({d}_{sdw}\). If the $p$-value \(p\) is less than 0.05, we can reject the null hypothesis and conclude that \( \mathcal{D}_{pub} \) has been stolen. On the other hand, if the null hypothesis can't be rejected, we think the suspect is innocent. 

\section{Experiments}
\label{others}

We evaluate our method using six visual datasets (CIFAR10~\cite{krizhevsky2009learning}, CIFAR100~\cite{krizhevsky2009learning}, SVHN~\cite{netzer2011reading}, ImageNette~\cite{howard2019smaller}, ImageWoof~\cite{howard2019smaller} and ImageNet~\cite{deng2009imagenet}) and five contrastive learning algorithms (SimCLR, BYOL, SimSiam, MOCO v3, and DINO). ImageNette and ImageWoof are two non-overlapping subsets of ImageNet, each containing 10 classes.

The specific experimental setup is introduced in Section \ref{exp_setup}, results and analyses are presented in Section \ref{exp_res}, the application of our method on the ImageNet pre-trained models are demonstrated in Section \ref{exp-imagenet}, ablation studies are conducted in Section \ref{exp-ablation} and Appendix \ref{app_exp}. Specifically, Appendix \ref{app-ablation-NvNpvt} presents the ablation study of sampling size, the ablation study of global and local augmentation number is shown in Appendix \ref{app-ablation-mn}, 
and the ablation study of shadow dataset and hyperparameter \(a\) are featured in Appendix \ref{app-ablation-alphaDsdw}. The impact of shadow model's training hyperparameters is shown in Appendix \ref{app-ablation-sdw-model}. The anti-interference capability of our method is conducted in Section \ref{exp_anti_interference}, Appendix \ref{app-compare-bw} introduces the comparison with the method based on watermark. Appendix \ref{app-early} introduces the impact of early stopping. Appendix \ref{app-visual} presents some visualization results of our method.

\subsection{Experimental Setup}
\label{exp_setup}

For SimCLR, BYOL, SimSiam, and MoCo v3, we use VGG16~\cite{simonyan2014very}, and Resnet18~\cite{he2016deep} as encoder architectures. Additionally, we use ViT-T, ViT-S, and ViT-B~\cite{dosovitskiy2020image} for DINO. For \(\mathcal{M}_{sdw}\), we default to using ResNet18 and SimCLR as its encoder architecture and training algorithm.

To simulate \(\mathcal{D}_{alt}\), a dataset similar to \(\mathcal{D}_{pub}\) but without overlapping data (as described in Section \ref{problem_formulate}), we randomly divide a dataset into two subsets of equal size representing \(\mathcal{D}_{pub}\) and \(\mathcal{D}_{alt}\), respectively. For \(\mathcal{D}_{pvt}\), we set it as the testing set of the undivided dataset for convenience. Specific settings are as follows: 
\begin{itemize}[leftmargin=*]
    \item \textbf{Experiment 1}: \(\mathcal{D}_{pub}\) is random half of CIFAR10 training set and \(\mathcal{D}_{alt}\) is the other half. \( \mathcal{D}_{unre} \), \( \mathcal{D}_{sdw} \) and \( \mathcal{D}_{pvt} \) are SVHN, CIFAR100 and CIFAR10 testing set respectively.
    \item \textbf{Experiment 2}: \(\mathcal{D}_{pub}\) is random half of ImageNette training set and \(\mathcal{D}_{alt}\) is the other half. \( \mathcal{D}_{unre} \), \( \mathcal{D}_{sdw} \) and \( \mathcal{D}_{pvt} \) are ImageWoof, SVHN and ImageNette testing set respectively.
\end{itemize}

The settings for the remaining parameters are provided in Appendix \ref{app_exp_detail}. To simulate adversarial behavior, we pre-train \(\mathcal{M}_{sus}\) using \(\mathcal{D}_{pub}\), \(\mathcal{D}_{pub}\cup\mathcal{D}_{alt}\), \( \mathcal{D}_{unre} \), and \(\mathcal{D}_{alt}\), respectively, which is corresponds to the four cases in Section \ref{problem_formulate}.

Regarding evaluation metrics, in addition to using the $p$-value, we also use the sensitivity, specificity and AUROC. Sensitivity is the proportion of correctly predicted positive cases among all actual positive samples, and specificity is the proportion of correctly predicted negative cases among all actual negative samples. They reflect the ability to identify positive and negative samples, respectively.

When \(\mathcal{M}_{sus}\) pre-trained on \(\mathcal{D}_{pub}\) or \(\mathcal{D}_{pub}\cup\mathcal{D}_{alt}\), which means the suspect is illegal, \(p\) should be less than 0.05. When \(\mathcal{M}_{sus}\) pre-trained on \(\mathcal{D}_{alt}\) or \( \mathcal{D}_{unre} \), which means the suspect is legal, \(p\) should be greater than 0.05. 
We compare our method with two representative methods, as detailed below:
\begin{itemize}[leftmargin=*]
    \item \textbf{DI4SSL}~\cite{dziedzic2022dataset}: This is the most recent method for dataset inference targeting self-supervised encoders. It also applies to dataset ownership verification. The principle behind DI4SSL is that if the encoder is pre-trained on \(\mathcal{D}_{pub}\), the representations it outputs will have a higher log-likelihood on the defender's training data than on testing data. Conversely, if the encoder is not pre-trained on \(\mathcal{D}_{pub}\), this pattern will not be observed.
    \item \textbf{EncoderMI}~\cite{liu2021encodermi}: This is a classic method which designed for member inference on contrastive pre-trained models. The fundamental mechanism of EncoderMI is that the encoder produces similar representations for different augmentation of the training data. We have adapted this method to suit dataset ownership verification better. Specifically, we augment images from \(\mathcal{D}_{pub}\) and \(\mathcal{D}_{pvt}\) and input them into \(\mathcal{M}_{sus}\) and \(\mathcal{M}_{sdw}\). By comparing the distribution of  the output representations' similarity, we can determine potential dataset stealing. If \(\mathcal{M}_{sus}\) is pre-trained on \(\mathcal{D}_{pub}\), the representations' similarity of \(\mathcal{M}_{sus}\) will significantly exceed those of \(\mathcal{M}_{sdw}\), vice versa.
\end{itemize}

\subsection{Experimental Results}
\label{exp_res}

\begin{figure}[ht]
    \vspace{-1em}
    \centering
    \includegraphics[width=\textwidth, trim=4.0cm 1.0cm 4.5cm 13cm, clip]{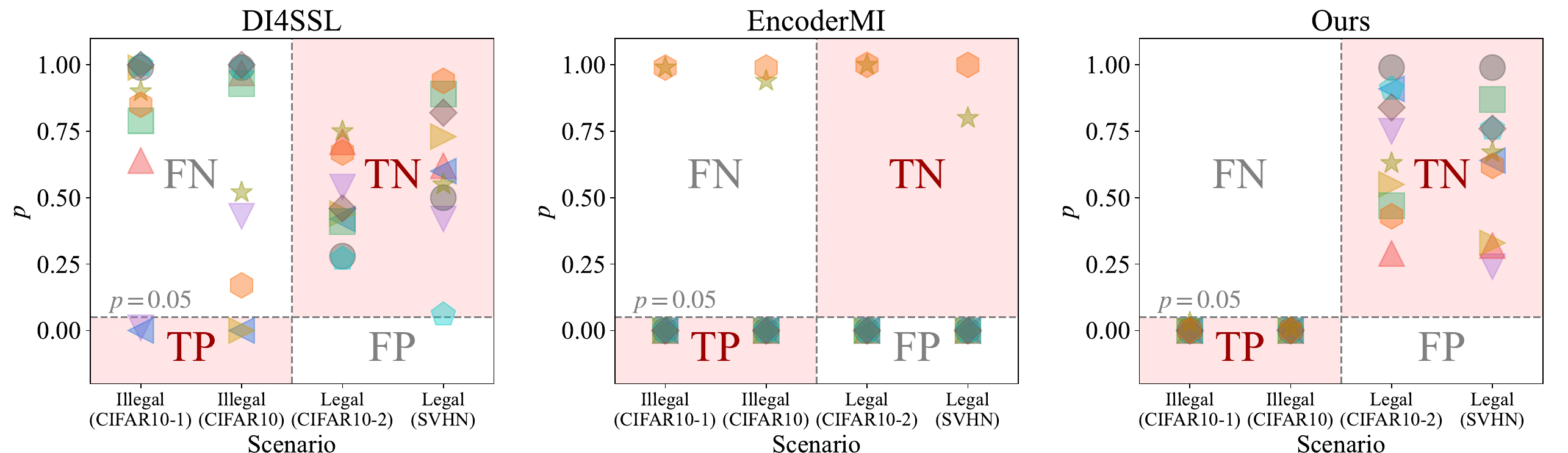}
    \includegraphics[width=\textwidth, trim=4.0cm 0.2cm 4.5cm 10.5cm, clip]{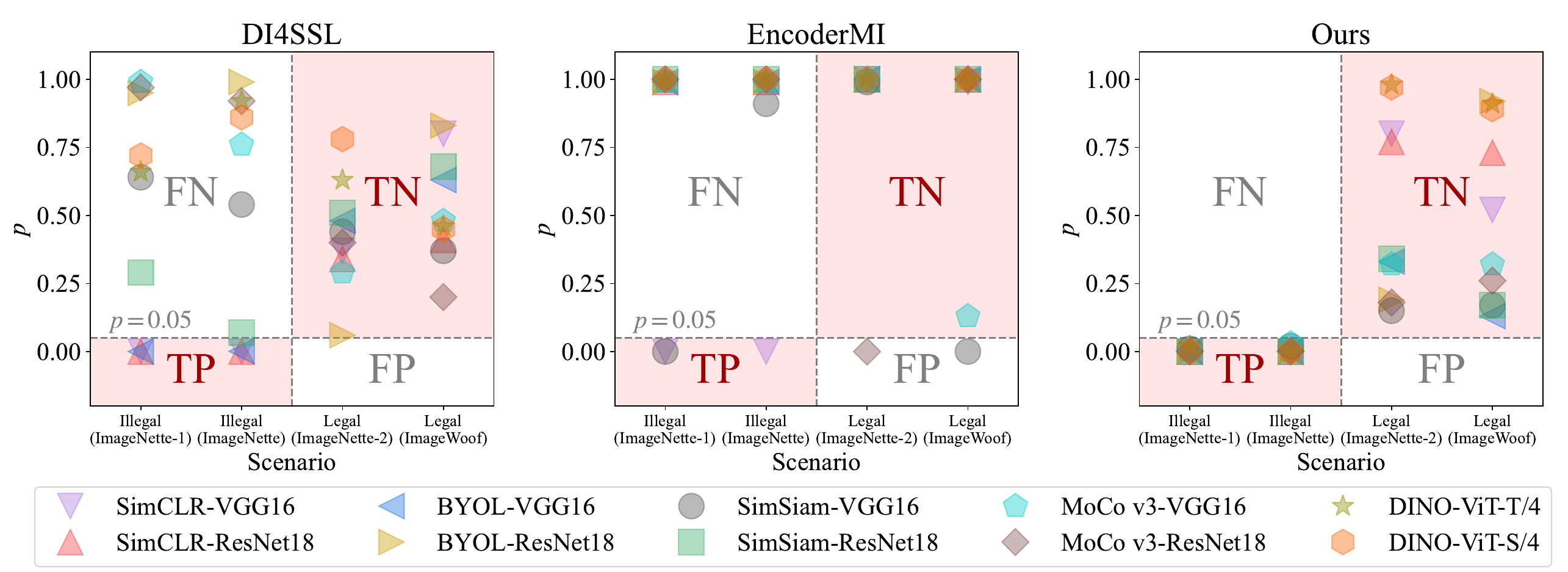}
    \caption{Experimental results of three methods on CIFAR10 (the first line) and ImageNette (the second line). Each value is an average of 3 trials. Each pattern represents a suspicious model trained using a specific architecture, contrastive learning method, and dataset. `SimCLR-VGG16' represents VGG16 trained using SimCLR, and the rest follows similarly. `CIFAR10-1' and `CIFAR10-2' are the two non-overlapping random halves of CIFAR10 training set, similarly for `ImageNette-1' and `ImageNette-2'. \(\mathcal{D}_{pub}\) is CIFAR10-1 and ImageNette-1 in two cases respectively. We consider illegal/legal behavior as positive/negative cases and classify each situation based on \(p\)-value. The datasets in parentheses on the x-axis are \(\mathcal{D}_{sus}\).}
    \vspace{-10pt}
    \label{fig:table1}
\end{figure}

Our approach is proven effective as illustrated in Figure \ref{fig:table1} (refer to Appendix \ref{app-p-CIFAR10} and \ref{app_imagenette} for specific \(p\)-values), which display the experimental results of baselines and our method on CIFAR10 and ImageNette. Note that when \(\mathcal{D}_{sus}\) is CIFAR10-1 (ImageNette-1) or CIFAR10 (ImageNette), \(\mathcal{D}_{sus}\) includes \(\mathcal{D}_{pub}\) (\(\mathcal{D}_{pub}\) is CIFAR10-1 and ImageNette-1 in two cases respectively), which implies the suspect is illegal, and \(p\) should be less than 0.05. However, when \(\mathcal{D}_{sus}\) is CIFAR10-2 (ImageNette-2) or SVHN (ImageWoof), the suspect did not use \(\mathcal{D}_{pub}\) and is legal, so \(p\) should be greater than 0.05.

The two baselines struggle to accurately distinguish the legality of various scenarios. There are a large number of false positive or false negative samples in all cases. In contrast, our method consistently produces correct results in all cases. Unlike the baselines, which model high-dimensional representations containing a large amount of redundant information directly, our method refines the most valuable information from these representations.

\begin{minipage}[t]{0.45\textwidth} 
    This crucial information, contrastive relationship gap, is extracted based on the characteristics of contrastive learning. Therefore, our method is not constrained by the encoder architecture and training algorithm, achieving desirable outcomes in various scenarios. As shown in Table \ref{auroc-table}, we calculate sensitivity , specificity and AUROC based on the experimental results on CIFAR10 and ImageNette, which demonstrates the superiority of our method quantitatively. 
\end{minipage}
\hfill
\begin{minipage}[t]{0.52\textwidth} 
    \vspace{-7pt}
    \centering
    \captionof{table}{Sensitivity, specificity, and AUROC of three methods on CIFAR10 and ImageNette.} 
    \label{auroc-table}
    \vspace{-2pt}
    \adjustbox{width=\textwidth}{
    \begin{tabular}{ccccc}
    \toprule
    Dataset&Method&Sensitivity&Specificity&AUROC\\ \hline
    
    \multirow{3}{*}{CIFAR10}&DI4SSL& 0.2&1.0&0.6\\
     &EncoderMI& 0.8&0.2& 0.5\\
     &Ours& \textbf{1.0}&\textbf{1.0}& \textbf{1.0}\\ \cline{1-5}
    
    \multirow{3}{*}{ImageNette}&DI4SSL& 0.3&1.0&0.775\\
     &EncoderMI& 0.15&0.9& 0.5\\
     &Ours& \textbf{1.0}&\textbf{1.0}& \textbf{1.0}\\
      \bottomrule
    \end{tabular}}
\end{minipage}
Sensitivity and specificity reflect the algorithm's ability to identify positive and negative samples.







\subsection{The Application of Our Method on ImageNet}
\label{exp-imagenet}

To validate the efficacy of our method in real-world scenarios, we conduct dataset ownership verification on ImageNet, a large-scale visual dataset containing over 14 million images across 1000 classes, using ten pre-trained encoders. The architecture of these encoders includes CNN and ViT, and they are pre-trained using the six popular contrastive learning methods currently. Among these, the pre-trained model for DINO is obtained from the official repository\footnote{\href{https://github.com/facebookresearch/dino}{https://github.com/facebookresearch/dino}}, while the models for the other contrastive learning methods are sourced from MMSelfSup\footnote{\href{https://mmselfsup.readthedocs.io/en/latest/model_zoo.html}{https://mmselfsup.readthedocs.io/en/latest/model\_zoo.html}}.
In our experiments, we designate \(\mathcal {D}_{pvt}\) as the validation set of ImageNet and \(\mathcal {D}_{sdw}\) as SVHN. The architecture and training algorithm of \(\mathcal {M}_{sdw}\) are ResNet18 and SimCLR, respectively. Parameter settings are provided in Appendix \ref{app_exp_detail}. As shown in Table \ref{imagenet-table}, the experimental outcomes demonstrate that our method is well-suited for pre-trained models on ImageNet, even when using only 0.1\% of ImageNet data for dataset ownership verification. Conversely, the performances of baselines are unsatisfactory.

\begin{table}[t]
\caption{The results (\(p\)-values) of baselines and our method applied on ImageNet. `\(\mathcal{D}_{sus}\)' is the dataset used to pre-train \(\mathcal{M}_{sus}\). Each value is an average of 3 trials. \(\mathcal {D}_{sus}\) and \(\mathcal {D}_{pub}\) are both ImageNet. Note that in this scenario, \(\mathcal {D}_{sus}\) includes \(\mathcal {D}_{pub}\), making the suspect's behavior illegal, and the \(p\)-values should be less than 0.05.}
\label{imagenet-table}
\centering

\begin{tabular}{ccccc}

\toprule
 {Method}& Model&   {DI4SSL}&  {EncoderMI}&{Ours}     \\ \hline

SimCLR& \multirow{4}{*}{ResNet50}&0.15&  1&\(10^{-3}\)\\  BYOL& &                  0.91&  1&              \(10^{-3}\)\\  SimSiam& &                  0.56&  1&              \(10^{-4}\)\\
  SwAV& &   0.88&  1& \(10^{-4}\)\\ \cline{1-2} 
  \multirow{3}{*}{MoCo v3}& ResNet50&                  0.51& 1&              \(10^{-3}\)\\
                          &      ViT-S/16&                  0.99&  \(10^{-159}\)&              \(10^{-4}\)\\
 &  ViT-B/16&   0.99&  \(10^{-158}\)& \(10^{-4}\)\\ \cline{1-2} \multirow{3}{*}{DINO}& ResNet50&                  0.99&  1&             \(10^{-4}\)\\
  & ViT-S/16&   0.99&  1& \(10^{-3}\)\\
                               & ViT-B/16&                  0.99&  1&              \(10^{-3}\)\\  
\bottomrule
\end{tabular}
\end{table}

\subsection{Ablation Studies}
\label{exp-ablation}

\subsubsection{The Impact of Multi-scale Augmentation in Unary and Binary Relationship}

We use pre-trained models on ImageNet to verify the effectiveness and robustness of unary and binary relationship's multi-scale augmentations. Specifically, the models are ResNet50 and ViT-B/16 pre-trained by DINO. Both \(\mathcal {D}_{pub}\) and \(\mathcal {D}_{sus}\) are ImageNet. As shown in Table \ref{ablation-1-table}, The combined use of unary and binary relationship's multi-scale augmentations outperform other choices. This superiority is attributed to its attempts to activate the encoder's contrastive relationship gap from various angles, thereby endowing it with strong generalization capabilities to adapt to different encoders.

\begin{table}[t]
\caption{The impact of multi-scale augmentation in unary and binary relationship. Both \(\mathcal {D}_{pub}\) and \(\mathcal {D}_{sus}\) are ImageNet. `DINO-ResNet50' represents ResNet50 trained using DINO, with `DINO-ViT-B/16' being similar. Note that the suspect is illegal in this case, and the \(p\)-values should be less than 0.05. \textbf{Bold} and \underline{underline} respectively represent the best and second best results.}
\label{ablation-1-table}
\centering
\adjustbox{width=\textwidth}{
\begin{tabular}{ccccccccc}

\toprule
{Study Subject}& \(S_U^{gg}\)&\(S_U^{gl}\)&\(S_U^{ll}\)& \(S_B^{gg}\)& \(S_B^{gl}\)&\(S_B^{ll}\)& {DINO-ResNet50} & {DINO-ViT-B/16}\\  \hline

\multirow{2}{*}{Unary/Binary Relationship}& \checkmark& \checkmark& \checkmark& & &&0.02 & 0.01\\ &  & & & \checkmark& \checkmark&\checkmark&                              \(3.1 \times {10^{ - 3}}\)&\(2.4\times10^{-3}\)\\ \cline{1-7} \multirow{6}{*}{Global/Local View}&  \checkmark& & & \checkmark& &&                              0.06 & 0.02\\ 
 &  & \checkmark& & & \checkmark&&  \(\mathbf{3.3\times10^{-4}}\)& \(2.5\times10^{-3}\)\\ & & & \checkmark& & &\checkmark&                              \(3.9\times10^{-3}\)& \(1.4\times10^{-3}\)\\
                          & \checkmark& \checkmark& & \checkmark& \checkmark&&                              \(8.2\times10^{-4}\)& \(2.5\times10^{-3}\)\\
 & & \checkmark& \checkmark& & \checkmark&\checkmark&  \(5.0\times10^{-4}\)& \(1.4\times10^{-3}\)\\ & \checkmark& & \checkmark& \checkmark& &\checkmark&                              \(1.8\times10^{-3}\)& \(\mathbf{1.0\times10^{-3}}\)\\ \cline{1-7}
 Ours& \checkmark& \checkmark& \checkmark& \checkmark& \checkmark&\checkmark&  \(\underline{4.2\times10^{-4}}\)& \(\underline{1.1\times10^{-3}}\)\\ 
\bottomrule
\end{tabular}}
\vspace{-10pt}
\end{table}

\subsubsection{The Impact of Sample Number of \texorpdfstring{\(\mathcal{D}_{pub}\)}{Lg} and \texorpdfstring{\(\mathcal{D}_{alt}\)}{Lg}}

We study the impact of the sample number of \(\mathcal{D}_{pub}\) and \(\mathcal{D}_{alt}\) on our method. Specifically, we denote the proportion of \(\mathcal{D}_{pub}\) in \(\mathcal{D}_{pub}\cup\mathcal{D}_{alt}\) as \(r\). And $\mathcal{D}_{pub}\cup\mathcal{D}_{alt}$ is always CIFAR10.
For example, when $r=0.1$, $\mathcal{D}_{pub}$ is 10\% of the CIFAR10 training set randomly sampled, while $\mathcal{D}_{alt}$ consists of the remaining 90\%. Similarly, when $r=0.2$, $\mathcal{D}_{pub}$ is 20\% of the CIFAR10 training set randomly sampled, and $\mathcal{D}_{alt}$ is the remaining 80\%.

\begin{wrapfigure}{r}{0.3\textwidth} 
  \includegraphics[width=\linewidth, trim=2cm 0.2cm 22cm 3cm, clip]{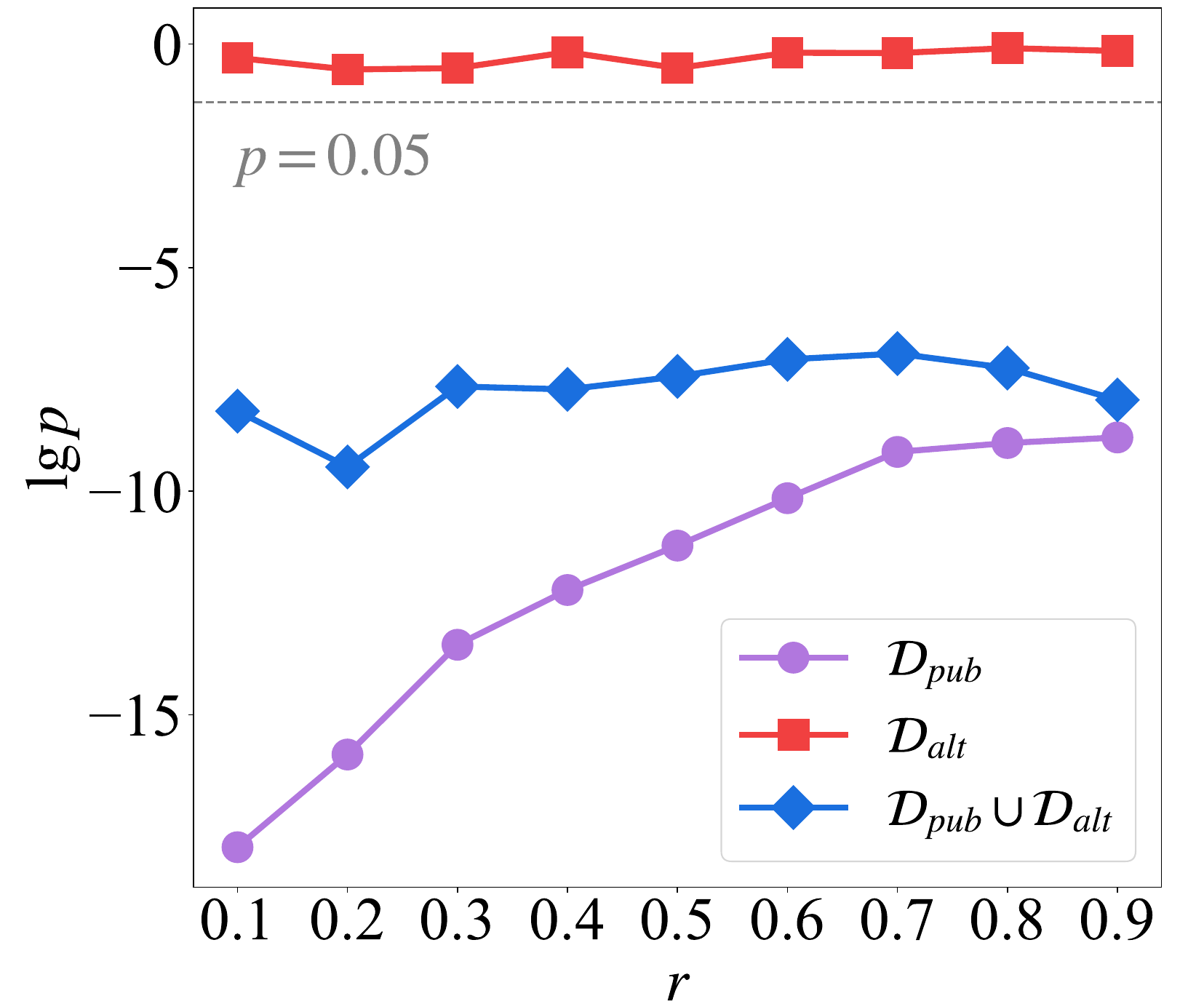}
  \caption{The impact of the ratio of \(\mathcal{D}_{pub}\) to \(\mathcal{D}_{pub}\cup\mathcal{D}_{alt}\) on our method. Each point is the $p$-value (log-transformed) of the model trained on the corresponding dataset.}
  \label{fig:ablation_r}
  \vspace{-40pt} 
\end{wrapfigure}

Then we use ResNet18 pre-trained using SimCLR to obverse the performance changes of our method under different \(r\) values. In Figure \ref{fig:ablation_r}, each point represents the $p$-value (log-transformed) of the model trained on the corresponding dataset. It shows our method demonstrates good robustness to the sample number of \(\mathcal{D}_{pub}\) and \(\mathcal{D}_{alt}\).

\subsection{The Anti-interference Capability of Our Method}
\label{exp_anti_interference}

\subsubsection{The Impact of Privacy Training Method}

Private training methods ~\cite{abadi2016deep, papernot2018scalable} are typically used to protect private, non-open-source datasets. In our scenario, the suspect might employ private training methods to obscure their illegal activities and interfere with the defender's dataset ownership verification, even if it reduces the encoder's normal performance. Therefore, we chose the classic private training method DP-SGD ~\cite{abadi2016deep} and conducted the following experiments. Specifically, we trained the suspicious encoder on ImageNette using DP-SGD or not. The \(\epsilon\) for DP-SGD is 50, and the maximum norm for gradient clipping is 1.2. 
The results are shown in Table \ref{tab:dpsgd-simclr} and Table \ref{tab:dpsgd-simsiam}, indicating that our method remains effective in this more arduous scenario.





\begin{table}[h]
    \centering
    \begin{minipage}{0.49\linewidth}
        \centering
        \caption{Results (\(p\)-values) on SimCLR.}
        \begin{tabular}{ccc}
            \toprule
            Model &{w/o DP-SGD}&{w/ DP-SGD} \\ \hline
             VGG16&\(10^{-27}\)&\(10^{-14}\)\\
             ResNet18&\(10^{-14}\)&\(10^{-13}\)\\
              \bottomrule
        \end{tabular}
        \label{tab:dpsgd-simclr}
    \end{minipage}%
    \hfill \hspace{2.5pt}
    \begin{minipage}{0.49\linewidth}
        \centering
        \caption{Results (\(p\)-values) on SimSiam.}
        \begin{tabular}{ccc}
            \toprule
            Model &{w/o DP-SGD}&{w/ DP-SGD} \\ \hline
             VGG16&\(0.01\)&\(0.01\) \\
             ResNet18&\(10^{-4}\)&\(10^{-3}\) \\
              \bottomrule
        \end{tabular}
        \label{tab:dpsgd-simsiam}
    \end{minipage}
\end{table}

\subsubsection{The Application of Our Method on Fine-tuned Encoders}
\label{exp_finetune}

We also challenge the scenario where \(\mathcal{M}_{sus}\) is applied to downstream tasks. Specifically, we train the entire classifier on CIFAR10 and CIFAR100 respectively, whose backbone is a ResNet50 pre-trained on ImageNet using SimCLR. Similarly, in the black-box environment, we can only use the predicted probability vectors of the input samples. The results are shown in Table \ref{tab:finetune-cifar10} and Table \ref{tab:finetune-cifar100}. `\( \mathcal D_{downstream} \)' is the dataset of downstream tasks. `Acc' represents the accuracy on downstream tasks.  \(\mathcal {D}_{sus}\) and \(\mathcal {D}_{pub}\) are both ImageNet. Note that in this scenario, \(\mathcal {D}_{sus}\) includes \(\mathcal {D}_{pub}\), making the suspect's behavior illegal, and the \(p\)-values should be less than 0.05. Moreover, we set the hyperparameter \(a\) to 5 to amplify the contrastive relationship gap. Excitingly, even after fine-tuning, we are still able to identify the suspect's theft. For details on fine-tuning, please refer to Appendix \ref{app-finetune}.

\vspace{-0.5em}
\begin{table}[h]
    \centering
    \begin{minipage}{0.49\linewidth}
        \centering
        \caption{Fine-tuning Results on CIFAR-10.}
        \begin{tabular}{cccc}
            \toprule
            \( \mathcal D_{downstream} \) & Epoch &{\(p\)(↓)} &Acc \\ \hline
             \multirow{4}{*}{CIFAR10}& 50&\(10^{-4}\)&0.87 \\
             & 100& \(10^{-3}\)&0.88 \\
             & 150&\(10^{-8}\)&0.88\\
             & 200&\(10^{-4}\)&0.89\\
              \bottomrule
        \end{tabular}
        \label{tab:finetune-cifar10}
    \end{minipage}%
    \hfill \hspace{2.5pt}
    \begin{minipage}{0.49\linewidth}
        \centering
        \caption{Fine-tuning Results on CIFAR-100.}
        \begin{tabular}{cccc}
            \toprule
            \( \mathcal D_{downstream} \) & Epoch &{\(p\)(↓)} &Acc \\ \hline
             \multirow{4}{*}{CIFAR100}& 50&\(10^{-6}\)&0.44 \\
             & 100& \(10^{-4}\)&0.50 \\
             & 150&\(10^{-5}\)&0.63\\
             & 200&\(10^{-3}\)&0.66\\
              \bottomrule
        \end{tabular}
        \label{tab:finetune-cifar100}
    \end{minipage}
\end{table}
\vspace{-0.5em}

\vspace{-0.5em}
\subsection{The Time Cost of Our Method}
\vspace{-0.5em}

\begin{minipage}[t]{0.5\textwidth} 
    We calculated the time required for our method and DI4SSL to perform a single verification on ImageNet. The experiments were conducted using an NVIDIA GeForce RTX 4090. The encoder is a ResNet50 pre-trained on ImageNet using SimCLR. As shown in Table \ref{cost-table}, the time consumption of our method is significantly less than that of DI4SSL. 
\end{minipage}
\hfill
\begin{minipage}[t]{0.45\textwidth} 
    \vspace{-13pt}
    \centering
    \captionof{table}{The time required for our method and DI4SSL to execute once on ImageNet.} 
    \label{cost-table}
    \begin{tabular}{ccc}
    \toprule
    {Method}& {Time Consumption}&{\(p\)}(↓)\\  \hline
    
    DI4SSL&10014s&1\\
    Ours&293s&\(10^{-3}\)\\
     \bottomrule
    \end{tabular}
\end{minipage}
This is because our method only requires inferring on a small subset of ImageNet (depending on \(k_{pub}\), \(k_{pvt}\), \(M\) and \(N\)), whereas DI4SSL needs to infer the entire dataset. Additionally, our method was properly validated (\(p<0.05\)), further demonstrating its superiority.






\vspace{-0.5em}
\subsection{Limitations}
\label{exp_limit}
\vspace{-0.5em}

Not all encoders are pre-trained using contrastive learning. Masked Image Modeling (MIM)~\cite{girdhar2023omnimae,he2022masked} is also a significant method for pre-training encoders. However, as shown in Appendix \ref{app-mae}, our method doesn't effectively apply to encoders pre-trained via MIM. This is because that the representations learned through MIM are harder to distinguish compared to those from contrastive learning~\cite{zhou2022mimco}, although MIM-based pre-training methods demonstrate superior performance in downstream tasks. This results in less pronounced unary and binary relational gaps in the representations. We plan to refine this aspect in our future work.

\vspace{-0.5em}
\section{Conclusion}
\vspace{-0.5em}

High-quality open-source datasets are essential for the rapid development of deep learning. We propose a method for verifying dataset ownership in contrastive learning to protect the legitimate right of dataset owners. Specifically, we propose the concept of contrastive relationship gap based on the unary and binary relationship of contrastive pre-trained models. The experiment proves that it can effectively verify dataset ownership. Promising future work includes (1) extending our method to other self-supervised learning approaches; (2) adapting our method to protect other types of data (e.g., text); (3) exploring other privacy risks associated with encoders.

\section{Acknowledgements}

This work was partially supported by the Pioneer R\&D Program of Zhejiang (No.2024C01021), and Zhejiang Provincial Natural Science Foundation of China (LQ24F020020, LD24F020011).


\bibliography{iclr2025_conference}

\begin{thebibliography}{56}
\providecommand{\natexlab}[1]{#1}
\providecommand{\url}[1]{\texttt{#1}}
\expandafter\ifx\csname urlstyle\endcsname\relax
  \providecommand{\doi}[1]{doi: #1}\else
  \providecommand{\doi}{doi: \begingroup \urlstyle{rm}\Url}\fi

\bibitem[Abadi et~al.(2016)Abadi, Chu, Goodfellow, McMahan, Mironov, Talwar, and Zhang]{abadi2016deep}
Martin Abadi, Andy Chu, Ian Goodfellow, H~Brendan McMahan, Ilya Mironov, Kunal Talwar, and Li~Zhang.
\newblock Deep learning with differential privacy.
\newblock In \emph{Proceedings of the 2016 ACM SIGSAC conference on computer and communications security}, pp.\  308--318, 2016.

\bibitem[Albelwi(2022)]{albelwi2022survey}
Saleh Albelwi.
\newblock Survey on self-supervised learning: auxiliary pretext tasks and contrastive learning methods in imaging.
\newblock \emph{Entropy}, 24\penalty0 (4):\penalty0 551, 2022.

\bibitem[Boneh \& Franklin(2001)Boneh and Franklin]{boneh2001identity}
Dan Boneh and Matt Franklin.
\newblock Identity-based encryption from the weil pairing.
\newblock In \emph{Annual international cryptology conference}, pp.\  213--229. Springer, 2001.

\bibitem[Carlini et~al.(2022)Carlini, Chien, Nasr, Song, Terzis, and Tramer]{carlini2022membership}
Nicholas Carlini, Steve Chien, Milad Nasr, Shuang Song, Andreas Terzis, and Florian Tramer.
\newblock Membership inference attacks from first principles.
\newblock In \emph{2022 IEEE Symposium on Security and Privacy (SP)}, pp.\  1897--1914. IEEE, 2022.

\bibitem[Caron et~al.(2020)Caron, Misra, Mairal, Goyal, Bojanowski, and Joulin]{caron2020unsupervised}
Mathilde Caron, Ishan Misra, Julien Mairal, Priya Goyal, Piotr Bojanowski, and Armand Joulin.
\newblock Unsupervised learning of visual features by contrasting cluster assignments.
\newblock \emph{Advances in neural information processing systems}, 33:\penalty0 9912--9924, 2020.

\bibitem[Caron et~al.(2021)Caron, Touvron, Misra, J{\'e}gou, Mairal, Bojanowski, and Joulin]{caron2021emerging}
Mathilde Caron, Hugo Touvron, Ishan Misra, Herv{\'e} J{\'e}gou, Julien Mairal, Piotr Bojanowski, and Armand Joulin.
\newblock Emerging properties in self-supervised vision transformers.
\newblock In \emph{Proceedings of the IEEE/CVF international conference on computer vision}, pp.\  9650--9660, 2021.

\bibitem[Chen et~al.(2020)Chen, Kornblith, Norouzi, and Hinton]{chen2020simple}
Ting Chen, Simon Kornblith, Mohammad Norouzi, and Geoffrey Hinton.
\newblock A simple framework for contrastive learning of visual representations.
\newblock In \emph{International conference on machine learning}, pp.\  1597--1607. PMLR, 2020.

\bibitem[Chen \& He(2021)Chen and He]{chen2021exploring}
Xinlei Chen and Kaiming He.
\newblock Exploring simple siamese representation learning.
\newblock In \emph{Proceedings of the IEEE/CVF conference on computer vision and pattern recognition}, pp.\  15750--15758, 2021.

\bibitem[Chen et~al.(2021{\natexlab{a}})Chen, Xie, and He]{chen2021empirical}
Xinlei Chen, Saining Xie, and Kaiming He.
\newblock An empirical study of training self-supervised vision transformers.
\newblock In \emph{Proceedings of the IEEE/CVF international conference on computer vision}, pp.\  9640--9649, 2021{\natexlab{a}}.

\bibitem[Chen et~al.(2021{\natexlab{b}})Chen, Wang, Bender, Ding, Jia, Li, and Song]{chen2021refit}
Xinyun Chen, Wenxiao Wang, Chris Bender, Yiming Ding, Ruoxi Jia, Bo~Li, and Dawn Song.
\newblock Refit: a unified watermark removal framework for deep learning systems with limited data.
\newblock In \emph{Proceedings of the 2021 ACM Asia Conference on Computer and Communications Security}, pp.\  321--335, 2021{\natexlab{b}}.

\bibitem[Choi et~al.(2024)Choi, Shavit, and Duvenaud]{choi2024tools}
Dami Choi, Yonadav Shavit, and David~K Duvenaud.
\newblock Tools for verifying neural models' training data.
\newblock \emph{Advances in Neural Information Processing Systems}, 36, 2024.

\bibitem[Choquette-Choo et~al.(2021)Choquette-Choo, Tramer, Carlini, and Papernot]{choquette2021label}
Christopher~A Choquette-Choo, Florian Tramer, Nicholas Carlini, and Nicolas Papernot.
\newblock Label-only membership inference attacks.
\newblock In \emph{International conference on machine learning}, pp.\  1964--1974. PMLR, 2021.

\bibitem[Cox et~al.(2002)Cox, Miller, Bloom, and Honsinger]{cox2002digital}
Ingemar Cox, Matthew Miller, Jeffrey Bloom, and Chris Honsinger.
\newblock Digital watermarking.
\newblock \emph{Journal of Electronic Imaging}, 11\penalty0 (3):\penalty0 414--414, 2002.

\bibitem[Deng et~al.(2009)Deng, Dong, Socher, Li, Li, and Fei-Fei]{deng2009imagenet}
Jia Deng, Wei Dong, Richard Socher, Li-Jia Li, Kai Li, and Li~Fei-Fei.
\newblock Imagenet: A large-scale hierarchical image database.
\newblock In \emph{2009 IEEE conference on computer vision and pattern recognition}, pp.\  248--255. Ieee, 2009.

\bibitem[Dosovitskiy et~al.(2020)Dosovitskiy, Beyer, Kolesnikov, Weissenborn, Zhai, Unterthiner, Dehghani, Minderer, Heigold, Gelly, et~al.]{dosovitskiy2020image}
Alexey Dosovitskiy, Lucas Beyer, Alexander Kolesnikov, Dirk Weissenborn, Xiaohua Zhai, Thomas Unterthiner, Mostafa Dehghani, Matthias Minderer, Georg Heigold, Sylvain Gelly, et~al.
\newblock An image is worth 16x16 words: Transformers for image recognition at scale.
\newblock \emph{arXiv preprint arXiv:2010.11929}, 2020.

\bibitem[Dwork(2006)]{dwork2006differential}
Cynthia Dwork.
\newblock Differential privacy.
\newblock In \emph{International colloquium on automata, languages, and programming}, pp.\  1--12. Springer, 2006.

\bibitem[Dziedzic et~al.(2022)Dziedzic, Duan, Kaleem, Dhawan, Guan, Cattan, Boenisch, and Papernot]{dziedzic2022dataset}
Adam Dziedzic, Haonan Duan, Muhammad~Ahmad Kaleem, Nikita Dhawan, Jonas Guan, Yannis Cattan, Franziska Boenisch, and Nicolas Papernot.
\newblock Dataset inference for self-supervised models.
\newblock \emph{Advances in Neural Information Processing Systems}, 35:\penalty0 12058--12070, 2022.

\bibitem[Fang et~al.(2023)Fang, Jia, Thudi, Yaghini, Choquette-Choo, Dullerud, Chandrasekaran, and Papernot]{fang2023proof}
Congyu Fang, Hengrui Jia, Anvith Thudi, Mohammad Yaghini, Christopher~A Choquette-Choo, Natalie Dullerud, Varun Chandrasekaran, and Nicolas Papernot.
\newblock Proof-of-learning is currently more broken than you think.
\newblock In \emph{2023 IEEE 8th European Symposium on Security and Privacy (EuroS\&P)}, pp.\  797--816. IEEE, 2023.

\bibitem[Gao et~al.(2022)Gao, Zhuang, Lin, Cheng, Sun, Li, and Shen]{gao2022disco}
Yuting Gao, Jia-Xin Zhuang, Shaohui Lin, Hao Cheng, Xing Sun, Ke~Li, and Chunhua Shen.
\newblock Disco: Remedying self-supervised learning on lightweight models with distilled contrastive learning.
\newblock In \emph{European Conference on Computer Vision}, pp.\  237--253. Springer, 2022.

\bibitem[Girdhar et~al.(2023)Girdhar, El-Nouby, Singh, Alwala, Joulin, and Misra]{girdhar2023omnimae}
Rohit Girdhar, Alaaeldin El-Nouby, Mannat Singh, Kalyan~Vasudev Alwala, Armand Joulin, and Ishan Misra.
\newblock Omnimae: Single model masked pretraining on images and videos.
\newblock In \emph{Proceedings of the IEEE/CVF conference on computer vision and pattern recognition}, pp.\  10406--10417, 2023.

\bibitem[Grill et~al.(2020)Grill, Strub, Altch{\'e}, Tallec, Richemond, Buchatskaya, Doersch, Avila~Pires, Guo, Gheshlaghi~Azar, et~al.]{grill2020bootstrap}
Jean-Bastien Grill, Florian Strub, Florent Altch{\'e}, Corentin Tallec, Pierre Richemond, Elena Buchatskaya, Carl Doersch, Bernardo Avila~Pires, Zhaohan Guo, Mohammad Gheshlaghi~Azar, et~al.
\newblock Bootstrap your own latent-a new approach to self-supervised learning.
\newblock \emph{Advances in neural information processing systems}, 33:\penalty0 21271--21284, 2020.

\bibitem[Guo et~al.(2023)Guo, Li, Wang, Xia, Huang, Liu, and Li]{guo2023domain}
Junfeng Guo, Yiming Li, Lixu Wang, Shu-Tao Xia, Heng Huang, Cong Liu, and Bo~Li.
\newblock Domain watermark: Effective and harmless dataset copyright protection is closed at hand.
\newblock \emph{Advances in Neural Information Processing Systems}, 36, 2023.

\bibitem[Hayase et~al.(2021)Hayase, Kong, Somani, and Oh]{hayase2021spectre}
Jonathan Hayase, Weihao Kong, Raghav Somani, and Sewoong Oh.
\newblock Spectre: Defending against backdoor attacks using robust statistics.
\newblock In \emph{International Conference on Machine Learning}, pp.\  4129--4139. PMLR, 2021.

\bibitem[He et~al.(2016)He, Zhang, Ren, and Sun]{he2016deep}
Kaiming He, Xiangyu Zhang, Shaoqing Ren, and Jian Sun.
\newblock Deep residual learning for image recognition.
\newblock In \emph{Proceedings of the IEEE conference on computer vision and pattern recognition}, pp.\  770--778, 2016.

\bibitem[He et~al.(2020)He, Fan, Wu, Xie, and Girshick]{he2020momentum}
Kaiming He, Haoqi Fan, Yuxin Wu, Saining Xie, and Ross Girshick.
\newblock Momentum contrast for unsupervised visual representation learning.
\newblock In \emph{Proceedings of the IEEE/CVF conference on computer vision and pattern recognition}, pp.\  9729--9738, 2020.

\bibitem[He et~al.(2022)He, Chen, Xie, Li, Doll{\'a}r, and Girshick]{he2022masked}
Kaiming He, Xinlei Chen, Saining Xie, Yanghao Li, Piotr Doll{\'a}r, and Ross Girshick.
\newblock Masked autoencoders are scalable vision learners.
\newblock In \emph{Proceedings of the IEEE/CVF conference on computer vision and pattern recognition}, pp.\  16000--16009, 2022.

\bibitem[Hogg et~al.(2013)Hogg, McKean, Craig, et~al.]{hogg2013introduction}
Robert~V Hogg, Joseph~W McKean, Allen~T Craig, et~al.
\newblock \emph{Introduction to mathematical statistics}.
\newblock Pearson Education India, 2013.

\bibitem[Howard(2019)]{howard2019smaller}
Jeremy Howard.
\newblock A smaller subset of 10 easily classified classes from imagenet, and a little more french.
\newblock \emph{URL https://github.com/fastai/imagenette}, 2019.

\bibitem[Hu et~al.(2022)Hu, Salcic, Sun, Dobbie, Yu, and Zhang]{hu2022membership}
Hongsheng Hu, Zoran Salcic, Lichao Sun, Gillian Dobbie, Philip~S Yu, and Xuyun Zhang.
\newblock Membership inference attacks on machine learning: A survey.
\newblock \emph{ACM Computing Surveys (CSUR)}, 54\penalty0 (11s):\penalty0 1--37, 2022.

\bibitem[Jia et~al.(2021)Jia, Yaghini, Choquette-Choo, Dullerud, Thudi, Chandrasekaran, and Papernot]{jia2021proof}
Hengrui Jia, Mohammad Yaghini, Christopher~A Choquette-Choo, Natalie Dullerud, Anvith Thudi, Varun Chandrasekaran, and Nicolas Papernot.
\newblock Proof-of-learning: Definitions and practice.
\newblock In \emph{2021 IEEE Symposium on Security and Privacy (SP)}, pp.\  1039--1056. IEEE, 2021.

\bibitem[Kadian et~al.(2021)Kadian, Arora, and Arora]{kadian2021robust}
Poonam Kadian, Shiafali~M Arora, and Nidhi Arora.
\newblock Robust digital watermarking techniques for copyright protection of digital data: A survey.
\newblock \emph{Wireless Personal Communications}, 118:\penalty0 3225--3249, 2021.

\bibitem[Karimi \& Tang(2020)Karimi and Tang]{karimi2020decision}
Hamid Karimi and Jiliang Tang.
\newblock Decision boundary of deep neural networks: Challenges and opportunities.
\newblock In \emph{Proceedings of the 13th International Conference on Web Search and Data Mining}, pp.\  919--920, 2020.

\bibitem[Karimi et~al.(2019)Karimi, Derr, and Tang]{karimi2019characterizing}
Hamid Karimi, Tyler Derr, and Jiliang Tang.
\newblock Characterizing the decision boundary of deep neural networks.
\newblock \emph{arXiv preprint arXiv:1912.11460}, 2019.

\bibitem[Khamitkar()]{khamitkarsurvey}
Siddhi Khamitkar.
\newblock A survey on fully homomorphic encryption.
\newblock \emph{IOSR Journal of Computer Engineering (IOSR-JCE) e-ISSN}, pp.\  2278--0661.

\bibitem[Krizhevsky et~al.(2009)Krizhevsky, Hinton, et~al.]{krizhevsky2009learning}
Alex Krizhevsky, Geoffrey Hinton, et~al.
\newblock Learning multiple layers of features from tiny images.
\newblock 2009.

\bibitem[Kwon(2021)]{kwon2021defending}
Hyun Kwon.
\newblock Defending deep neural networks against backdoor attack by using de-trigger autoencoder.
\newblock \emph{IEEE Access}, 2021.

\bibitem[Li et~al.(2023{\natexlab{a}})Li, Pang, Xi, Du, Ji, Yao, and Wang]{li2023embarrassingly}
Changjiang Li, Ren Pang, Zhaohan Xi, Tianyu Du, Shouling Ji, Yuan Yao, and Ting Wang.
\newblock An embarrassingly simple backdoor attack on self-supervised learning.
\newblock In \emph{Proceedings of the IEEE/CVF International Conference on Computer Vision}, pp.\  4367--4378, 2023{\natexlab{a}}.

\bibitem[Li et~al.(2022)Li, Bai, Jiang, Yang, Xia, and Li]{li2022untargeted}
Yiming Li, Yang Bai, Yong Jiang, Yong Yang, Shu-Tao Xia, and Bo~Li.
\newblock Untargeted backdoor watermark: Towards harmless and stealthy dataset copyright protection.
\newblock \emph{Advances in Neural Information Processing Systems}, 35:\penalty0 13238--13250, 2022.

\bibitem[Li et~al.(2023{\natexlab{b}})Li, Zhu, Yang, Jiang, Wei, and Xia]{li2023black}
Yiming Li, Mingyan Zhu, Xue Yang, Yong Jiang, Tao Wei, and Shu-Tao Xia.
\newblock Black-box dataset ownership verification via backdoor watermarking.
\newblock \emph{IEEE Transactions on Information Forensics and Security}, 2023{\natexlab{b}}.

\bibitem[Li et~al.(2018)Li, Ding, and Gao]{li2018decision}
Yu~Li, Lizhong Ding, and Xin Gao.
\newblock On the decision boundary of deep neural networks.
\newblock \emph{arXiv preprint arXiv:1808.05385}, 2018.

\bibitem[Liu et~al.(2021{\natexlab{a}})Liu, Jia, Qu, and Gong]{liu2021encodermi}
Hongbin Liu, Jinyuan Jia, Wenjie Qu, and Neil~Zhenqiang Gong.
\newblock Encodermi: Membership inference against pre-trained encoders in contrastive learning.
\newblock In \emph{Proceedings of the 2021 ACM SIGSAC Conference on Computer and Communications Security}, pp.\  2081--2095, 2021{\natexlab{a}}.

\bibitem[Liu et~al.(2021{\natexlab{b}})Liu, Zhu, and Bai]{liu2021wdnet}
Yang Liu, Zhen Zhu, and Xiang Bai.
\newblock Wdnet: Watermark-decomposition network for visible watermark removal.
\newblock In \emph{Proceedings of the IEEE/CVF Winter Conference on Applications of Computer Vision}, pp.\  3685--3693, 2021{\natexlab{b}}.

\bibitem[Liu et~al.(2022)Liu, Jia, Liu, and Gong]{liu2022stolenencoder}
Yupei Liu, Jinyuan Jia, Hongbin Liu, and Neil~Zhenqiang Gong.
\newblock Stolenencoder: stealing pre-trained encoders in self-supervised learning.
\newblock In \emph{Proceedings of the 2022 ACM SIGSAC Conference on Computer and Communications Security}, pp.\  2115--2128, 2022.

\bibitem[Maini et~al.(2021)Maini, Yaghini, and Papernot]{maini2021dataset}
Pratyush Maini, Mohammad Yaghini, and Nicolas Papernot.
\newblock Dataset inference: Ownership resolution in machine learning.
\newblock \emph{arXiv preprint arXiv:2104.10706}, 2021.

\bibitem[Netzer et~al.(2011)Netzer, Wang, Coates, Bissacco, Wu, Ng, et~al.]{netzer2011reading}
Yuval Netzer, Tao Wang, Adam Coates, Alessandro Bissacco, Baolin Wu, Andrew~Y Ng, et~al.
\newblock Reading digits in natural images with unsupervised feature learning.
\newblock In \emph{NIPS workshop on deep learning and unsupervised feature learning}, volume 2011, pp.\ ~7. Granada, Spain, 2011.

\bibitem[Papernot et~al.(2018)Papernot, Song, Mironov, Raghunathan, Talwar, and Erlingsson]{papernot2018scalable}
Nicolas Papernot, Shuang Song, Ilya Mironov, Ananth Raghunathan, Kunal Talwar, and {\'U}lfar Erlingsson.
\newblock Scalable private learning with pate.
\newblock \emph{arXiv preprint arXiv:1802.08908}, 2018.

\bibitem[Podilchuk \& Delp(2001)Podilchuk and Delp]{podilchuk2001digital}
Christine~I Podilchuk and Edward~J Delp.
\newblock Digital watermarking: algorithms and applications.
\newblock \emph{IEEE signal processing Magazine}, 18\penalty0 (4):\penalty0 33--46, 2001.

\bibitem[Sanyal et~al.(2022)Sanyal, Addepalli, and Babu]{sanyal2022towards}
Sunandini Sanyal, Sravanti Addepalli, and R~Venkatesh Babu.
\newblock Towards data-free model stealing in a hard label setting.
\newblock In \emph{Proceedings of the IEEE/CVF conference on computer vision and pattern recognition}, pp.\  15284--15293, 2022.

\bibitem[Sha et~al.(2023)Sha, He, Yu, Backes, and Zhang]{sha2023can}
Zeyang Sha, Xinlei He, Ning Yu, Michael Backes, and Yang Zhang.
\newblock Can't steal? cont-steal! contrastive stealing attacks against image encoders.
\newblock In \emph{Proceedings of the IEEE/CVF Conference on Computer Vision and Pattern Recognition}, pp.\  16373--16383, 2023.

\bibitem[Shen et~al.(2022)Shen, He, Han, and Zhang]{shen2022model}
Yun Shen, Xinlei He, Yufei Han, and Yang Zhang.
\newblock Model stealing attacks against inductive graph neural networks.
\newblock In \emph{2022 IEEE Symposium on Security and Privacy (SP)}, pp.\  1175--1192. IEEE, 2022.

\bibitem[Shokri et~al.(2017)Shokri, Stronati, Song, and Shmatikov]{shokri2017membership}
Reza Shokri, Marco Stronati, Congzheng Song, and Vitaly Shmatikov.
\newblock Membership inference attacks against machine learning models.
\newblock In \emph{2017 IEEE symposium on security and privacy (SP)}, pp.\  3--18. IEEE, 2017.

\bibitem[Simonyan \& Zisserman(2014)Simonyan and Zisserman]{simonyan2014very}
Karen Simonyan and Andrew Zisserman.
\newblock Very deep convolutional networks for large-scale image recognition.
\newblock \emph{arXiv preprint arXiv:1409.1556}, 2014.

\bibitem[Sun et~al.(2023)Sun, Li, Meng, Ao, Lyu, Li, and Zhang]{sun2023defending}
Xiaofei Sun, Xiaoya Li, Yuxian Meng, Xiang Ao, Lingjuan Lyu, Jiwei Li, and Tianwei Zhang.
\newblock Defending against backdoor attacks in natural language generation.
\newblock In \emph{Proceedings of the AAAI Conference on Artificial Intelligence}, volume~37, pp.\  5257--5265, 2023.

\bibitem[Tang et~al.(2023)Tang, Feng, Liu, Yang, and Hu]{tang2023did}
Ruixiang Tang, Qizhang Feng, Ninghao Liu, Fan Yang, and Xia Hu.
\newblock Did you train on my dataset? towards public dataset protection with cleanlabel backdoor watermarking.
\newblock \emph{ACM SIGKDD Explorations Newsletter}, 25\penalty0 (1):\penalty0 43--53, 2023.

\bibitem[Zhao et~al.(2024)Zhao, Fang, Wang, and Zhou]{zhao2024proof}
Zishuo Zhao, Zhixuan Fang, Xuechao Wang, and Yuan Zhou.
\newblock Proof-of-learning with incentive security.
\newblock \emph{arXiv preprint arXiv:2404.09005}, 2024.

\bibitem[Zhou et~al.(2022)Zhou, Yu, Luo, Wang, and Li]{zhou2022mimco}
Qiang Zhou, Chaohui Yu, Hao Luo, Zhibin Wang, and Hao Li.
\newblock Mimco: Masked image modeling pre-training with contrastive teacher.
\newblock In \emph{Proceedings of the 30th ACM International Conference on Multimedia}, pp.\  4487--4495, 2022.

\end{thebibliography}
\bibliographystyle{iclr2025_conference}


\clearpage
\appendix
\section*{Appendix}
\addcontentsline{toc}{section}{Appendix}
\renewcommand{\thesubsection}{\Alph{subsection}}

\subsection{The Details and Additional Supplements of Experiments}
\label{app_exp}

\subsubsection{Datasets Used}
\label{app_dataset}

\textbf{CIFAR10}~\cite{krizhevsky2009learning}: The CIFAR10 dataset consists of 32x32 colored images with 10 classes. There are 50000 training images and 10000 test images.

\textbf{CIFAR100}~\cite{krizhevsky2009learning}: The CIFAR100 dataset consists of 32x32 coloured images with 100 classes. There are 50000 training images and 10000 test images.

\textbf{SVHN}~\cite{netzer2011reading}: The SVHN dataset contains 32x32 coloured images with 10 classes. There are roughly 73000 training images, 26000 test images and 530000 "extra" images.

\textbf{ImageNette}~\cite{howard2019smaller}: ImageNette is a subset of 10 easily classified classes from Imagenet. It includes the following categories: tench, English springer, cassette player, chain saw, church, French horn, garbage truck, gas pump, golf ball and parachute. There are roughly 10000 training images and 4000 test images.

\textbf{ImageWoof}~\cite{howard2019smaller}: ImageWoof is a subset of 10 classes from Imagenet that aren't so easy to classify. It includes the following categories: Australian terrier, Border terrier, Samoyed, Beagle, Shih-Tzu, English foxhound, Rhodesian ridgeback, Dingo, Golden retriever, Old English sheepdog. There are approximately 9000 training images and 4000 test images.

\textbf{ImageNet}~\cite{deng2009imagenet}: Larger sized coloured images with 1000 classes. There are approximately 1 million training images and 50000 test images. As is commonly done, we resize all images to be of size 224x224.

\subsubsection{Experimental Details}
\label{app_exp_detail}

The ResNet18 trained on CIFAR10/CIFAR100 uses a convolutional kernel size of 3x3 with a stride of 1, instead of the default 7x7, and doesn't use a max pooling layer.

On CIFAR10/CIFAR100/SVHN, we pre-train the encoder for 800 epochs with a batch size of 512. On ImageNette/ImageWoof, the encoder with non-ViT-S/16 architecture is pre-trained for 800 epochs, while ViT-S/16 architecture is pre-trained for 2000 epochs with a batch size of 64. The initial learning rate for all pre-training sessions is set at 0.06 and adjusted using a Cosine Annealing scheduler. The optimizer is SGD, with a momentum of 0.9 and a weight decay of \(5 \times {10^{ - 4}}\). All experiments are conducted on four NVIDIA RTX A6000s and one NVIDIA GeForce RTX 4090. In all experiments, we set \(M = 2\) and \(N = 6\). Both \( T^g \) and \( T^l \) are composed of random cropping, color jitter, random flipping, and random grayscale, with respective cropping ranges of (0.4, 1.0) and (0.05, 0.4).

Furthermore, the settings for other parameters are as follows: 

\textbf{Experiment on CIFAR10.} We set \({k_{pub}} = 256\), \({k_{pvt}} = 128\), \(K = 30\) and \(a = 10000\).

\textbf{Experiment on ImageNette.} We set \({k_{pub}} = {k_{pvt}} = 32\), \(K = 50\) and \(a = 0.1\).

\textbf{Experiment on ImageNet.} We set \({k_{pub}} = {k_{pvt}} = 32\) and \(K = 50\) and \(a = 1\).

\subsubsection{Detailed Experimental Results on CIFAR10}
\label{app-p-CIFAR10}

This section presents the experimental results (\(p\)-values) of several baselines and our method on CIFAR10. \(\mathcal{D}_{pub}\) is CIFAR10-1. The results are shown in Table \ref{cifar10-table-DI4SSL}, Table \ref{cifar10-table-EncoderMI}, and Table \ref{cifar10-table-Ours}, respectively. `\(\mathcal{D}_{sus}\)' is the dataset used to pre-train \(\mathcal{M}_{sus}\). `CIFAR10-1' and `CIFAR10-2' are the two non-overlapping halves of CIFAR10 training set after a random split. Each value is an average of 3 trials. Note that when \(\mathcal{D}_{sus}\) is CIFAR10-1 or CIFAR10, the suspect used \(\mathcal{D}_{pub}\), and this scenario is illegal, so \(p\) should be less than 0.05. However, when \(\mathcal{D}_{sus}\) is CIFAR10-2 or SVHN, the suspect did not use \(\mathcal{D}_{pub}\), and this scenario is legal, so \(p\) should be greater than 0.05. 

\subsubsection{Detailed Experimental Results on ImageNette}
\label{app_imagenette}

This section presents the experimental results (\(p\)-values) of several baselines and our method on ImageNette. \(\mathcal{D}_{pub}\) is ImageNette-1. The results are shown in Table \ref{imagenette-table-DI4SSL}, Table \ref{imagenette-table-EncoderMI}, and Table \ref{imagenette-table-Ours}, respectively. `ImageNette-1' and `ImageNette-2' are the two non-overlapping random halves of ImageNette training set. Each value is an average of 3 trials. Note that when \(\mathcal{D}_{sus}\) is ImageNette-1 or ImageNette, the suspect is illegal, so \(p\) should be less than 0.05. However, when \(\mathcal{D}_{sus}\) is ImageNette-2 or SVHN, the suspect is legal, so \(p\) should be greater than 0.05. 

\subsubsection{The Impact of Shadow Model's Training Hyperparameter}
\label{app-ablation-sdw-model}

In the real world, the training hyperparameters of shadow models and suspicious models are often different. We analyzed whether these differences would affect our method. Specifically, we set different batch size (32 for the shadow model and 64 for the suspicious model), learning rate (0.01 for the shadow model and 0.06 for the suspicious model), and weight decay (1e-4 for the shadow model and 5e-4 for the suspicious model) for the shadow model compared to the suspicious model. As shown in Table \ref{sdw-para-table}, our method demonstrates good robustness to the training hyperparameter settings of the shadow model.

\begin{table}[ht]
\caption{The impact of shadow model's training hyperparameters on our method. Model is ResNet18. Both \(\mathcal{D}_{pub}\) and \(\mathcal{D}_{sus}\) are ImageNette. \(\mathcal{D}_{sdw}\) is SVHN.}
\label{sdw-para-table}
\centering
\adjustbox{width=\textwidth}{
\begin{tabular}{ccccc}

\toprule
{Method}& {All Same}&{Different Batch Size}&{Different Learning Rate}&{Different Weight Decay}\\  \hline

SimCLR& \(10^{-11}\) & \(10^{-6}\)& \(10^{-10}\)& \(10^{-5}\)\\
BYOL& \(10^{-10}\) & \(10^{-4}\)& \(10^{-3}\)& \(10^{-4}\)\\
SimSiam& \(10^{-5}\) & \(10^{-4}\)& \(10^{-3}\)& \(10^{-3}\)\\
 \bottomrule
\end{tabular}}
\end{table}

\subsubsection{The Details of Fine-tuning the Pre-trained Model}
\label{app-finetune}
We fine-tuned the encoder on CIFAR10/CIFAR100 using a learning rate of 0.001, a batch size of 512, a weight decay of 5e-4, and the SGD optimizer with a momentum of 0.9. The pre-trained ResNet50 on ImageNet is sourced from MMSelfSup\footnote{\href{https://mmselfsup.readthedocs.io/en/latest/model_zoo.html}{https://mmselfsup.readthedocs.io/en/latest/model\_zoo.html}}.

\clearpage
\begin{table}[ht]
\caption{\(p\)-values for each scenario of DI4SSL on CIFAR10.}
\label{cifar10-table-DI4SSL}
\centering
\begin{tabular}{>{\centering\arraybackslash}p{1.5cm}cccccc}

\toprule
\multirow{2}{*}{Alg} & \multirow{2}{*}{Method} &  \multirow{2}{*}{Model} 
                          & \multicolumn{4}{c} {\(\mathcal{D}_{sus}\)} \\ \cline{4-7} & & & {CIFAR10-1} & {CIFAR10} & {CIFAR10-2} & {SVHN} \\  \hline
                          
\multirow{10}{*}{DI4SSL}& \multirow{2}{*}{SimCLR}    & VGG16                & 0.01& 0.43 &              0.54&                        0.42\\
                          &     & ResNet18                  &                               0.64&                             0.97&                               0.71&                          0.62\\ \cline{2-3}

& \multirow{2}{*}{BYOL} & VGG16                &                               \(10^{-4}\)&                             \(10^{-16}\)&                               0.42&                          0.60\\
                          &     & ResNet18                  &                               0.99&                             \(10^{-5}\)&                              0.44&                          0.73\\ \cline{2-3}
& \multirow{2}{*}{SimSiam}& VGG16                &                               0.99&                             0.99&                               0.28&                          0.50\\
                          &     & ResNet18                  &                               0.79&                             0.93&                               0.41&                          0.89\\ \cline{2-3}

& \multirow{2}{*}{MoCo v3}& VGG16                &                               1&                            0.99&                               0.27&                         0.06\\
                          &     & ResNet18                  &                               1&                             1&                              0.46&                         0.82\\ \cline{2-3}
                          
& \multirow{2}{*}{DINO} & ViT-T/4&                               0.90&                             0.52&                               0.75&                          0.55\\
                          &     & ViT-S/4&                               0.85&                             0.17&                               0.67&                          0.94\\ 
\bottomrule
\end{tabular}
\vspace{-5pt}
\end{table}

\begin{table}[ht]
\caption{\(p\)-values for each scenario of EncoderMI on CIFAR10.}
\label{cifar10-table-EncoderMI}
\centering
\begin{tabular}{>{\centering\arraybackslash}p{1.5cm}cccccc}

\toprule
\multirow{2}{*}{Alg} & \multirow{2}{*}{Method} &  \multirow{2}{*}{Model} 
                          & \multicolumn{4}{c} {\(\mathcal{D}_{sus}\)} \\ \cline{4-7} & & & {CIFAR10-1} & {CIFAR10} & {CIFAR10-2} & {SVHN} \\  \hline

\multirow{10}{*}{EncoderMI}& \multirow{2}{*}{SimCLR}    & VGG16                & 0& 0&                               0&                          0\\
                          &     & ResNet18                  &                               0&                             0&                               0&                         0\\ \cline{2-3}

& \multirow{2}{*}{BYOL} & VGG16                &                               0&                             0&                               0&                         0\\
                          &     & ResNet18                  &                               0&                             0&                               0&                          \(10^{-43}\)\\ \cline{2-3}
& \multirow{2}{*}{SimSiam}& VGG16                &                               0&                             0&                               0&                          0\\
                          &     & ResNet18                  &                               0&                             0&                              \(10^{-13}\)&                         0\\ \cline{2-3}

& \multirow{2}{*}{MoCo v3}& VGG16                &                               0&                             0&                               0&                          0\\
                          &     & ResNet18                  &                               0&                             0&                               0&                         \(10^{-74}\)\\ \cline{2-3}
                          
& \multirow{2}{*}{DINO} & ViT-T/4&                              0.99&                             0.94&                              1&                          0.80\\
                          &     & ViT-S/4&                               0.99&                             0.99&                               1&                          1\\ 

\bottomrule
\end{tabular}
\vspace{-5pt}
\end{table}

\begin{table}[ht]
\caption{\(p\)-values for each scenario of our method on CIFAR10.}
\label{cifar10-table-Ours}
\centering
\begin{tabular}{>{\centering\arraybackslash}p{1.5cm}cccccc}

\toprule
\multirow{2}{*}{Alg} & \multirow{2}{*}{Method} &  \multirow{2}{*}{Model} 
                          & \multicolumn{4}{c} {\(\mathcal{D}_{sus}\)} \\ \cline{4-7} & & & {CIFAR10-1} & {CIFAR10} & {CIFAR10-2} & {SVHN} \\  \hline

\multirow{10}{*}{Ours}& \multirow{2}{*}{SimCLR}    & VGG16                & \(10^{-16}\)& \(10^{-12}\)&                               0.75&                          0.24\\
                          &     & ResNet18                  &                               \(10^{-12}\)&                             \(10^{-8}\)&                               0.29&                          0.32\\ \cline{2-3}

& \multirow{2}{*}{BYOL} & VGG16                &                               \(10^{-20}\)&                             \(10^{-18}\)&                               0.91&                          0.64\\
                          &     & ResNet18                  &                               \(10^{-17}\)&                             \(10^{-11}\)&                               0.55&                          0.33\\ \cline{2-3}
& \multirow{2}{*}{SimSiam}& VGG16                &                               \(10^{-4}\)&                            \(10^{-6}\)&                               0.99&                          0.99\\
                          &     & ResNet18                  &                               \(10^{-11}\)&                            \(10^{-5}\)&                              0.47&                          0.87\\ \cline{2-3}

& \multirow{2}{*}{MoCo v3}& VGG16                &                               \(10^{-11}\)&                             \(10^{-14}\)&                               0.91&                         0.76\\
                          &     & ResNet18                  &                               \(10^{-4}\)&                             \(10^{-3}\)&                               0.84&                          0.76\\ \cline{2-3}
                          
& \multirow{2}{*}{DINO} & ViT-T/4&                               0.03&                            0.01&                               0.63&                         0.67\\
                          &     & ViT-S/4&                               \(10^{-7}\)&                            \(10^{-7}\)&                              0.43&                          0.62\\  
\bottomrule
\end{tabular}
\end{table}


\begin{table}[ht]
\caption{\(p\)-values for each scenario of DI4SSL on ImageNette.}
\label{imagenette-table-DI4SSL}
\centering
\begin{tabular}{>{\centering\arraybackslash}p{1.5cm}cccccc}

\toprule
\multirow{2}{*}{Alg} & \multirow{2}{*}{Method} &  \multirow{2}{*}{Model} 
                          & \multicolumn{4}{c} {\(\mathcal{D}_{sus}\)} \\ \cline{4-7} & & & {ImageNette-1} & {ImageNette} & {ImageNette-2} & {ImageWoof} \\  \hline
                          
\multirow{10}{*}{DI4SSL}& \multirow{2}{*}{SimCLR}    & VGG16                & 0.01& 0.43 &              0.54&                        0.42\\
                          &     & ResNet18                  &                               0.64&                             0.97&                               0.71&                          0.62\\ \cline{2-3}

& \multirow{2}{*}{BYOL} & VGG16                &                               \(10^{-4}\)&                             \(10^{-16}\)&                               0.42&                          0.60\\
                          &     & ResNet18                  &                               0.99&                             \(10^{-5}\)&                              0.44&                          0.73\\ \cline{2-3}
& \multirow{2}{*}{SimSiam}& VGG16                &                               0.99&                             0.99&                               0.28&                          0.50\\
                          &     & ResNet18                  &                               0.79&                             0.93&                               0.41&                          0.89\\ \cline{2-3}

& \multirow{2}{*}{MoCo v3}& VGG16                &                               1&                            0.99&                               0.27&                         0.06\\
                          &     & ResNet18                  &                               1&                             1&                              0.46&                         0.82\\ \cline{2-3}
                          
& \multirow{2}{*}{DINO} & ViT-T/4&                               0.90&                             0.52&                               0.75&                          0.55\\
                          &     & ViT-S/4&                               0.85&                             0.17&                               0.67&                          0.94\\ 
\bottomrule
\end{tabular}
\vspace{-5pt}
\end{table}

\begin{table}[ht]
\caption{\(p\)-values for each scenario of EncoderMI on ImageNette.}
\label{imagenette-table-EncoderMI}
\centering
\begin{tabular}{>{\centering\arraybackslash}p{1.5cm}cccccc}

\toprule
\multirow{2}{*}{Alg} & \multirow{2}{*}{Method} &  \multirow{2}{*}{Model} 
                          & \multicolumn{4}{c} {\(\mathcal{D}_{sus}\)} \\ \cline{4-7} & & & {ImageNette-1} & {ImageNette} & {ImageNette-2} & {ImageWoof} \\  \hline

\multirow{10}{*}{EncoderMI}& \multirow{2}{*}{SimCLR}    & VGG16                & 0& 0&                               0&                          0\\
                          &     & ResNet18                  &                               0&                             0&                               0&                         0\\ \cline{2-3}

& \multirow{2}{*}{BYOL} & VGG16                &                               0&                             0&                               0&                         0\\
                          &     & ResNet18                  &                               0&                             0&                               0&                          \(10^{-43}\)\\ \cline{2-3}
& \multirow{2}{*}{SimSiam}& VGG16                &                               0&                             0&                               0&                          0\\
                          &     & ResNet18                  &                               0&                             0&                              \(10^{-13}\)&                         0\\ \cline{2-3}

& \multirow{2}{*}{MoCo v3}& VGG16                &                               0&                             0&                               0&                          0\\
                          &     & ResNet18                  &                               0&                             0&                               0&                         \(10^{-74}\)\\ \cline{2-3}
                          
& \multirow{2}{*}{DINO} & ViT-T/4&                              0.99&                             0.94&                              1&                          0.80\\
                          &     & ViT-S/4&                               0.99&                             0.99&                               1&                          1\\ 

\bottomrule
\end{tabular}
\vspace{-5pt}
\end{table}

\begin{table}[ht]
\caption{\(p\)-values for each scenario of our method on ImageNette.}
\label{imagenette-table-Ours}
\centering
\begin{tabular}{>{\centering\arraybackslash}p{1.5cm}cccccc}

\toprule
\multirow{2}{*}{Alg} & \multirow{2}{*}{Method} &  \multirow{2}{*}{Model} 
                          & \multicolumn{4}{c} {\(\mathcal{D}_{sus}\)} \\ \cline{4-7} & & & {ImageNette-1} & {ImageNette} & {ImageNette-2} & {ImageWoof} \\  \hline

\multirow{10}{*}{Ours}& \multirow{2}{*}{SimCLR}    & VGG16                & \(10^{-16}\)& \(10^{-12}\)&                               0.75&                          0.24\\
                          &     & ResNet18                  &                               \(10^{-12}\)&                             \(10^{-8}\)&                               0.29&                          0.32\\ \cline{2-3}

& \multirow{2}{*}{BYOL} & VGG16                &                               \(10^{-20}\)&                             \(10^{-18}\)&                               0.91&                          0.64\\
                          &     & ResNet18                  &                               \(10^{-17}\)&                             \(10^{-11}\)&                               0.55&                          0.33\\ \cline{2-3}
& \multirow{2}{*}{SimSiam}& VGG16                &                               \(10^{-4}\)&                            \(10^{-6}\)&                               0.99&                          0.99\\
                          &     & ResNet18                  &                               \(10^{-11}\)&                            \(10^{-5}\)&                              0.47&                          0.87\\ \cline{2-3}

& \multirow{2}{*}{MoCo v3}& VGG16                &                               \(10^{-11}\)&                             \(10^{-14}\)&                               0.91&                         0.76\\
                          &     & ResNet18                  &                               \(10^{-4}\)&                             \(10^{-3}\)&                               0.84&                          0.76\\ \cline{2-3}
                          
& \multirow{2}{*}{DINO} & ViT-T/4&                               0.03&                            0.01&                               0.63&                         0.67\\
                          &     & ViT-S/4&                               \(10^{-7}\)&                            \(10^{-7}\)&                              0.43&                          0.62\\  
\bottomrule
\end{tabular}
\end{table}

\clearpage
\subsubsection{The Impact of Sampling Size}
\label{app-ablation-NvNpvt}
\begin{figure}[ht]
    \centering
    \begin{subfigure}{0.49\textwidth}
        \centering
        \includegraphics[width=\textwidth, trim=15cm 0cm 7cm 0.5cm, clip]{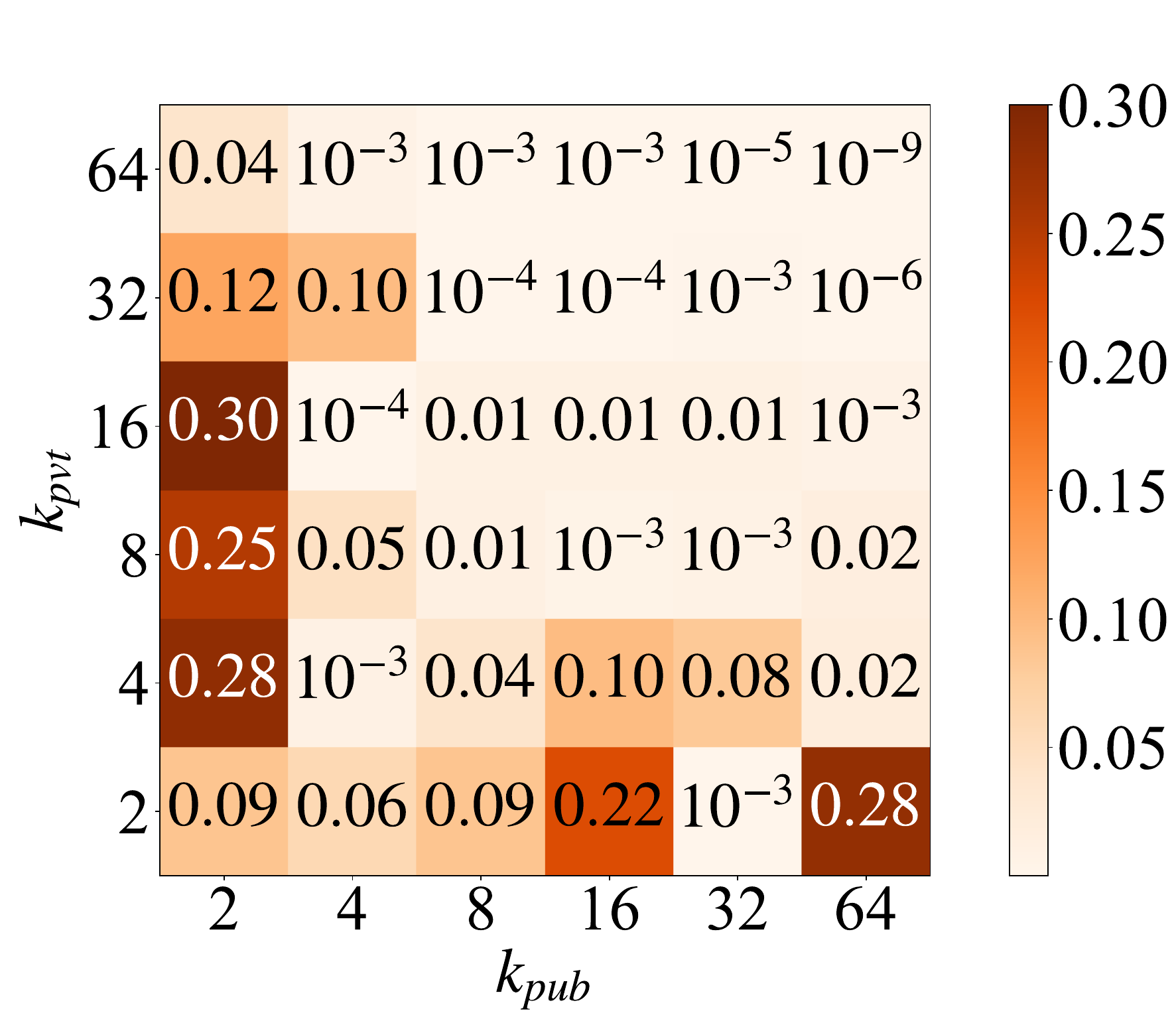}
        \caption{}
        \label{fig:ablation-NvNpvt-simclr}
    \end{subfigure}
    \hfill
    \begin{subfigure}{0.49\textwidth}
        \centering
        \includegraphics[width=\textwidth, trim=15cm 0cm 7cm 0.5cm, clip]{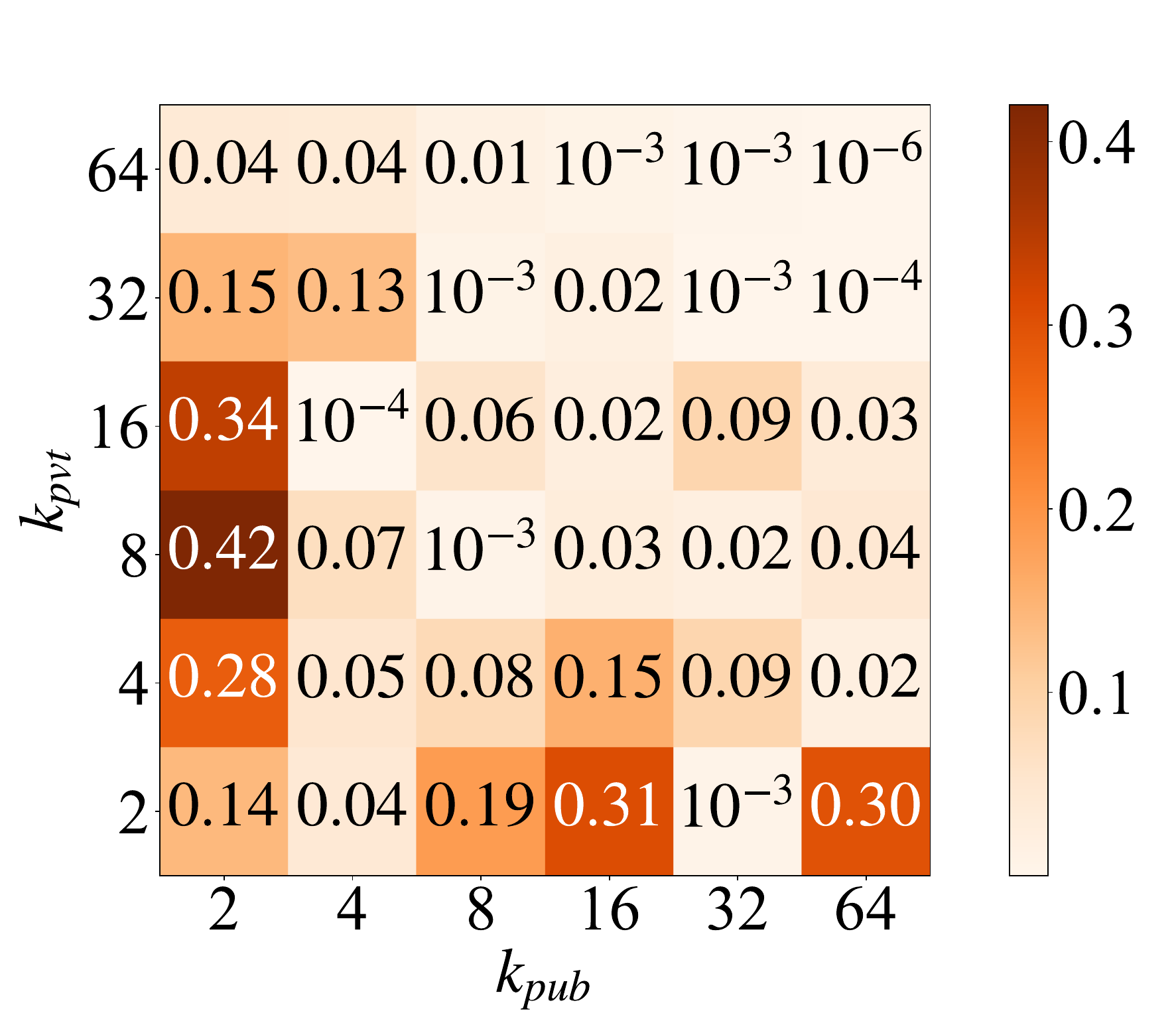}
        \caption{}
        \label{fig:ablation-NvNpvt-byol}
    \end{subfigure}
    \hfill
    \begin{subfigure}{0.49\textwidth}
        \centering
        \includegraphics[width=\textwidth, trim=15cm 0cm 7cm 0.5cm, clip]{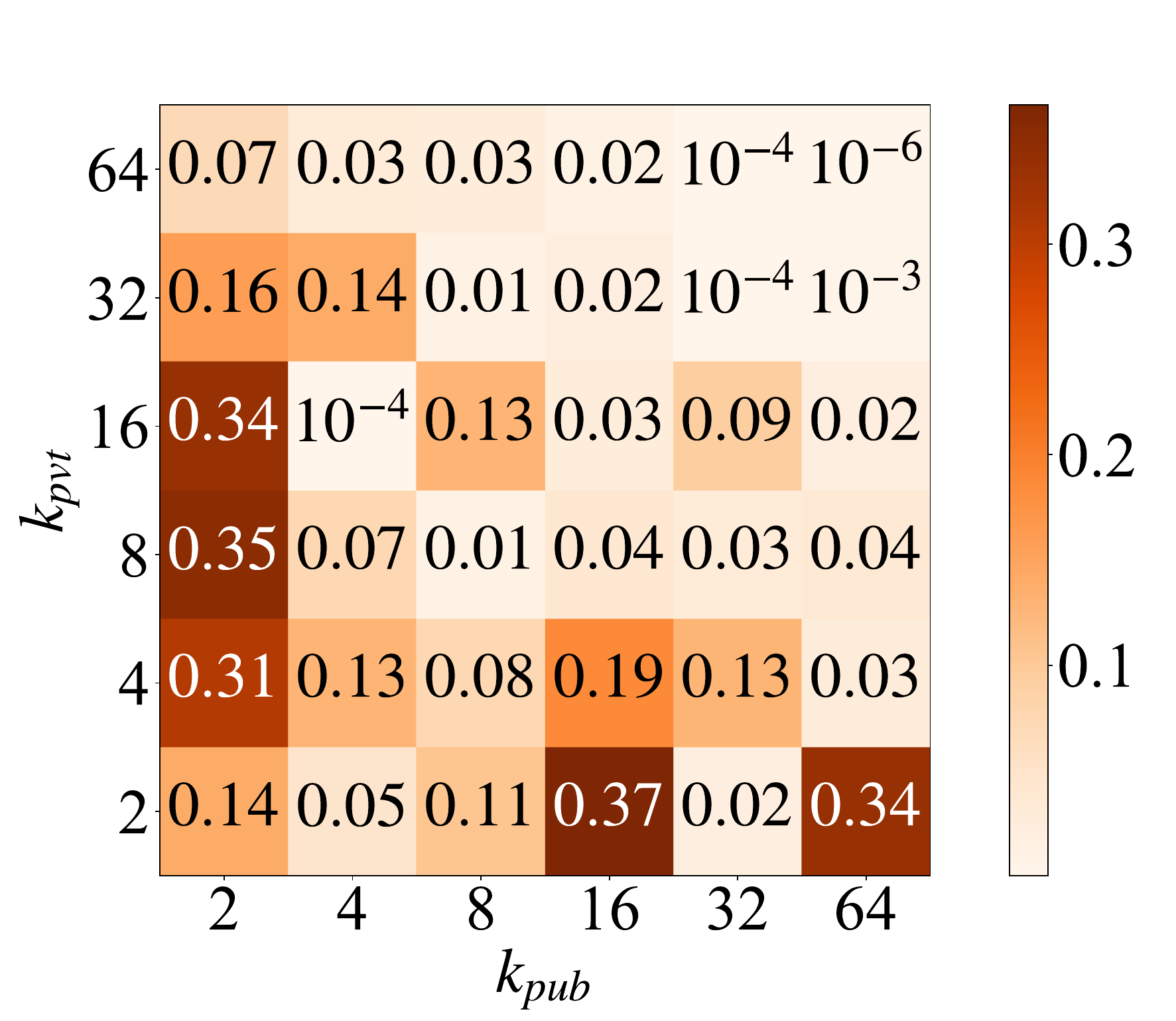}
        \caption{}
        \label{fig:ablation-NvNpvt-simsiam}
    \end{subfigure}
    \hfill
    \begin{subfigure}{0.49\textwidth}
        \centering
        \includegraphics[width=\textwidth, trim=15cm 0cm 7cm 0.5cm, clip]{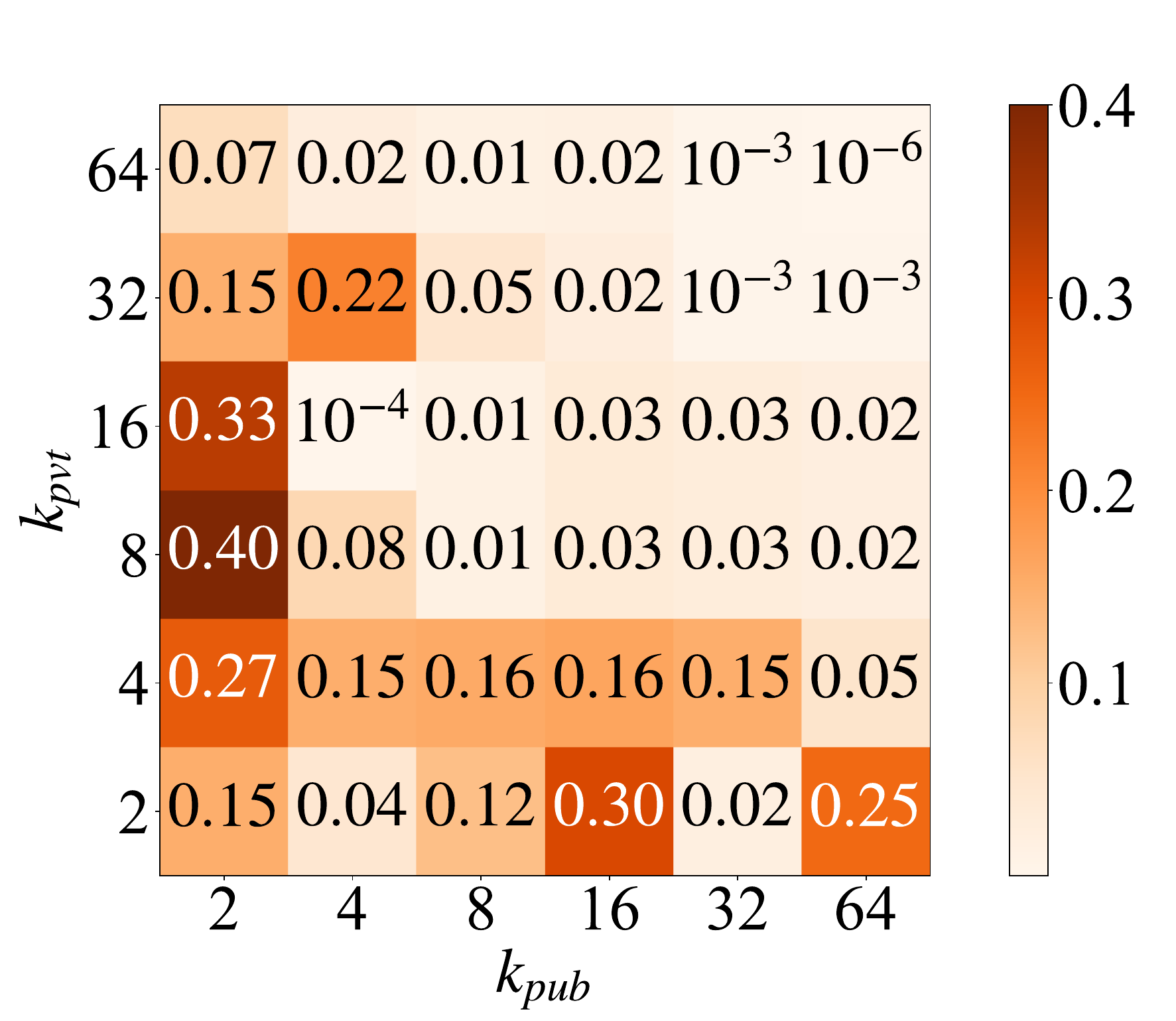}
        \caption{}
        \label{fig:ablation-NvNpvt-mocov3}
    \end{subfigure}
    \caption{The $p$-values obtained using pre-trained ResNet50 on ImageNet with different \(k_{pub}\) and \(k_{pvt}\) values. Each heatmap corresponds to the results of different training algorithms. Figure \ref{fig:ablation-NvNpvt-simclr}: SimCLR, Figure \ref{fig:ablation-NvNpvt-byol}: BYOL, Figure \ref{fig:ablation-NvNpvt-simsiam}: SimSiam, and Figure \ref{fig:ablation-NvNpvt-mocov3}: MoCo v3..}
    \label{fig:ablation-NvNpvt}
\end{figure}

We also evaluate the performance of our method using different amounts of data. Specifically, we conduct verification by selecting different sampling size \(k_{pub}\) and \(k_{pvt}\). We conduct experiments using pre-trained encoders on ImageNet, with the encoder architecture being ResNet50. Figure \ref{fig:ablation-NvNpvt} shows that, across various contrastive learning methods, the effectiveness of our method improves as \(k_{pub}\) and \(k_{pvt}\) increase. This is because larger sampling size better represent the distribution of the dataset, making the contrastive relationship gap of the encoder more pronounced. Note that in this scenario, \(\mathcal {D}_{sus}\) includes \(\mathcal {D}_{pub}\) are both ImageNet, and \(\mathcal {D}_{sus}\) includes \(\mathcal {D}_{pub}\), so the \(p\)-values should be less than 0.05.

\clearpage
\subsubsection{The Impact of Global and Local Augmentation Number}
\label{app-ablation-mn}
\begin{figure}[ht]
    \centering
    \begin{subfigure}{0.49\textwidth}
        \centering
        \includegraphics[width=\textwidth, trim=14cm 7cm 5cm 2cm, clip]{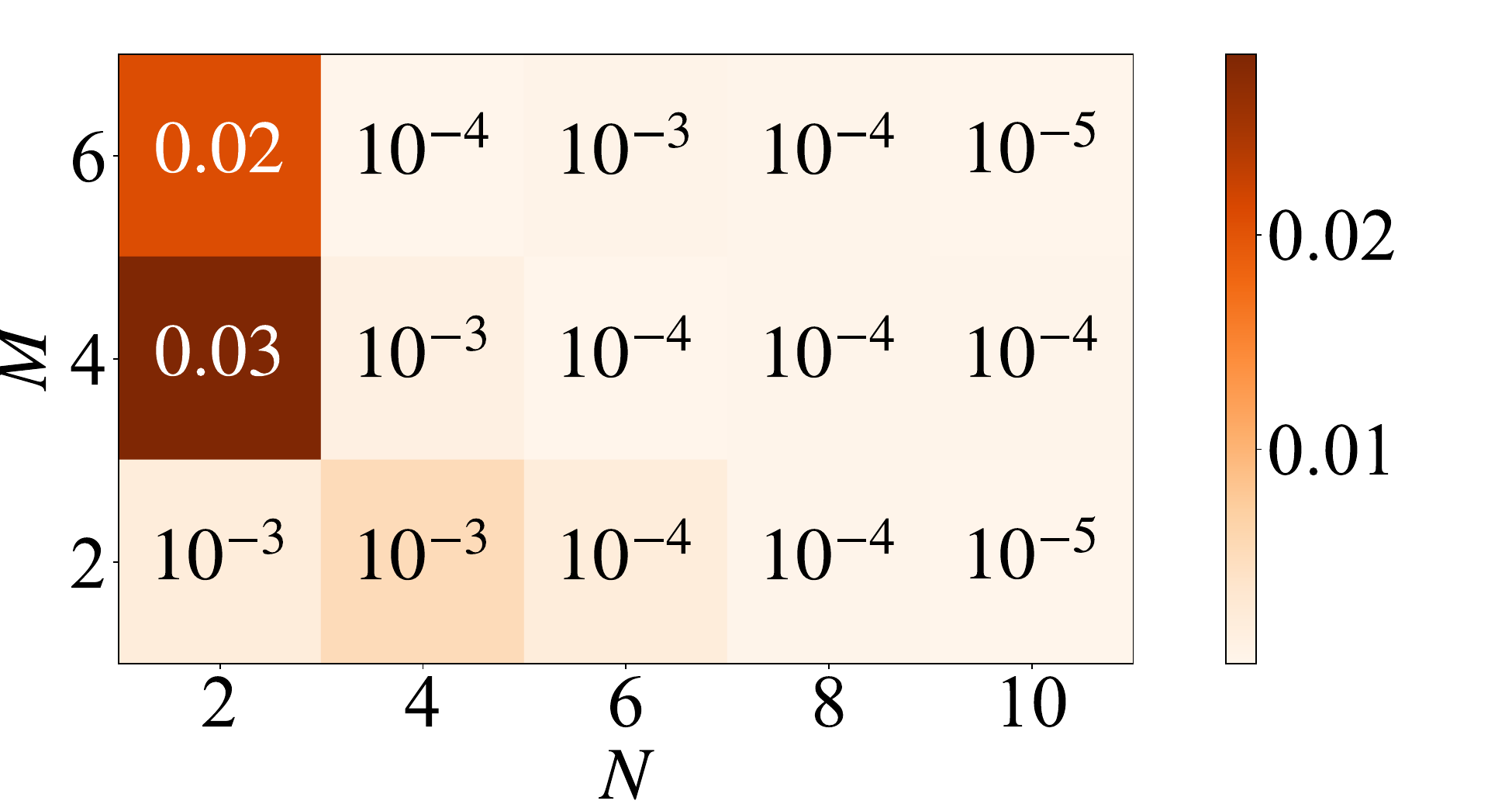}
        \caption{}
        \label{fig:ablation-mn-simclr}
    \end{subfigure}
    \hfill
    \begin{subfigure}{0.49\textwidth}
        \centering
        \includegraphics[width=\textwidth, trim=14cm 7cm 5cm 2cm, clip]{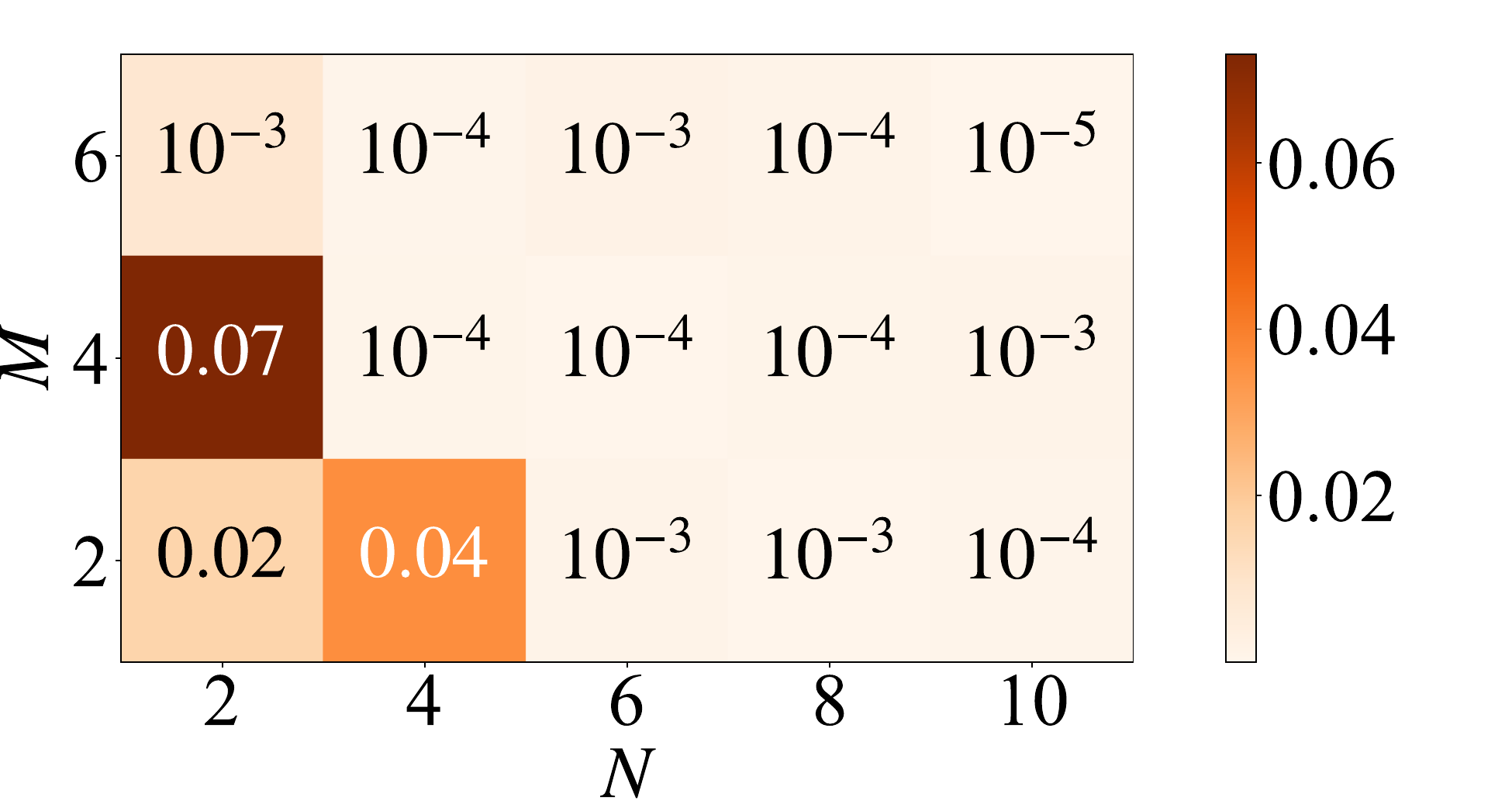}
        \caption{}
        \label{fig:ablation-mn-byol}
    \end{subfigure}
    \hfill
    \begin{subfigure}{0.49\textwidth}
        \centering
        \includegraphics[width=\textwidth, trim=14cm 7cm 5cm 2cm, clip]{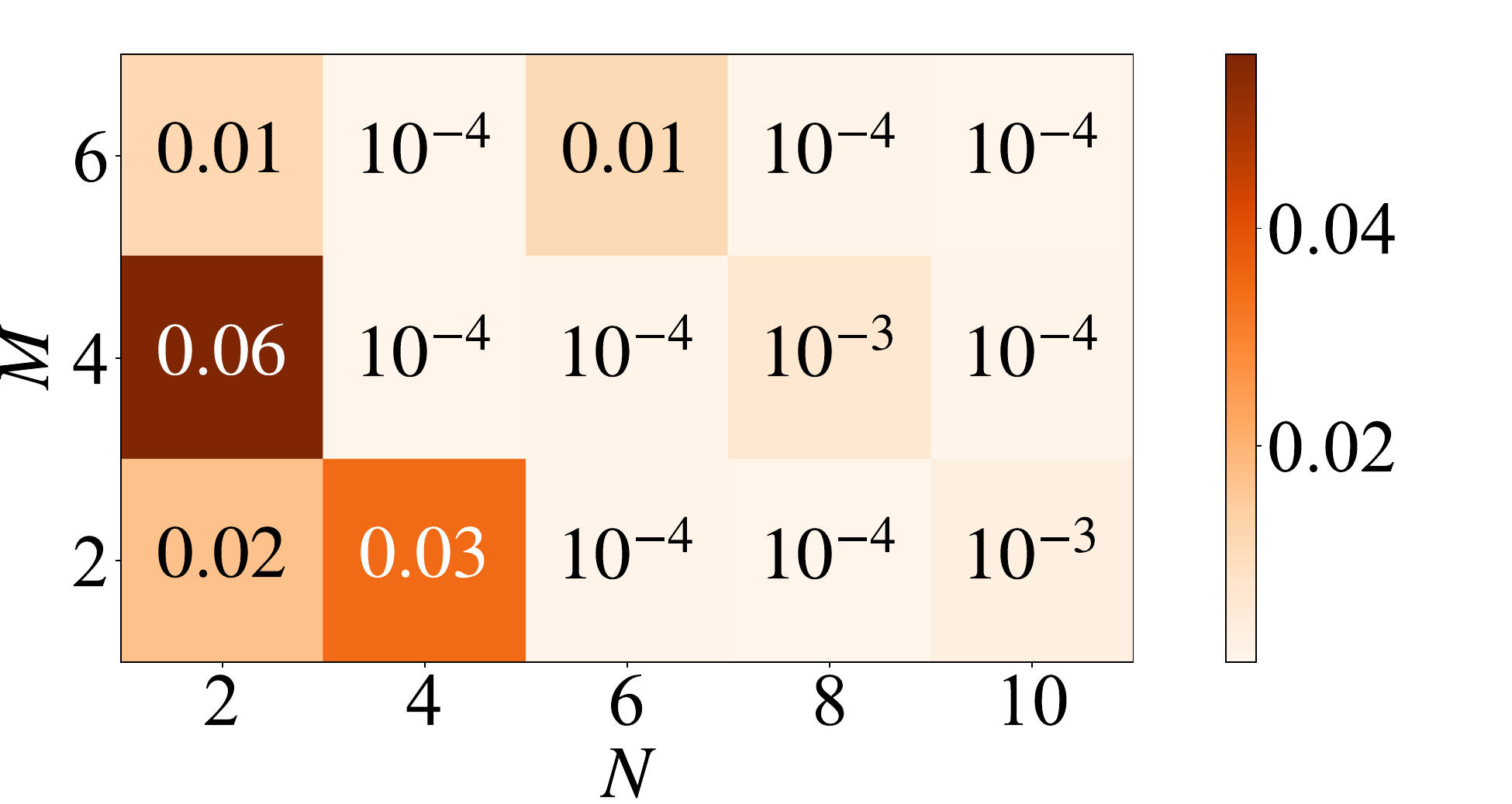}
        \caption{}
        \label{fig:ablation-mn-simsiam}
    \end{subfigure}
    \hfill
    \begin{subfigure}{0.49\textwidth}
        \centering
        \includegraphics[width=\textwidth, trim=14cm 7cm 5cm 2cm, clip]{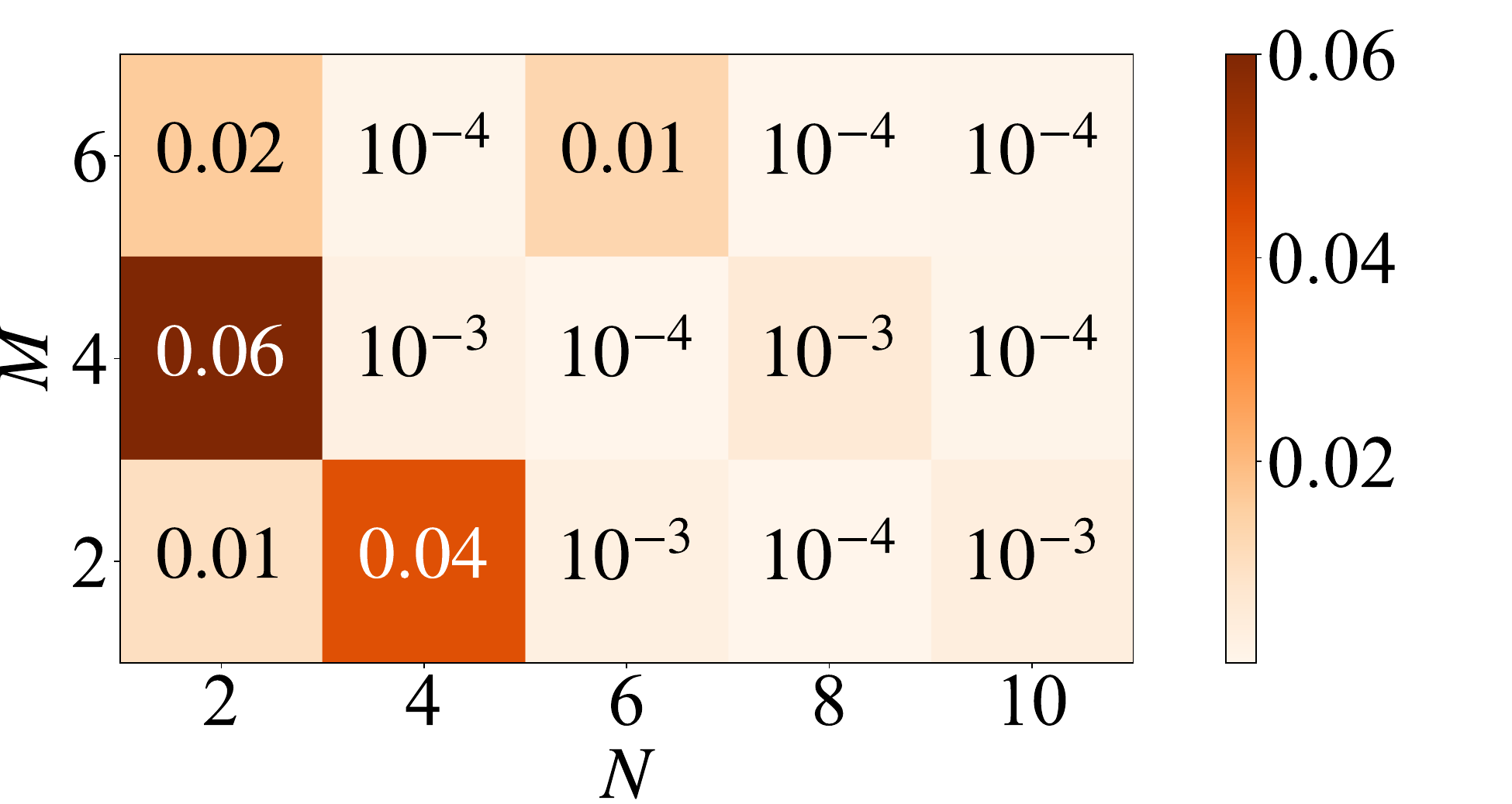}
        \caption{}
        \label{fig:ablation-mn-mocov3}
    \end{subfigure}
    \caption{The $p$-values obtained using pre-trained ResNet50 on ImageNet with different \(M\) and \(N\) values. Each heatmap corresponds to the results of different training algorithms. Figure \ref{fig:ablation-mn-simclr}: SimCLR, Figure \ref{fig:ablation-mn-byol}: BYOL, Figure \ref{fig:ablation-mn-simsiam}: SimSiam, and Figure \ref{fig:ablation-mn-mocov3}: MoCo v3.}
    \label{fig:ablation-mn}
\end{figure}

We evaluate the effectiveness of our method under different numbers of global augmentations \(M\) and local augmentations \(N\). We conduct experiments using pre-trained ResNet50 on ImageNet. Figure \ref{fig:ablation-mn} shows that, the performance of our method improves as \(M\) and \(N\) increase. This is because a greater number of augmentations provides more information to the encoder, thereby amplifying the contrastive relationship gap. Note that in this scenario, \(\mathcal {D}_{sus}\) includes \(\mathcal {D}_{pub}\) are both ImageNet, and \(\mathcal {D}_{sus}\) includes \(\mathcal {D}_{pub}\), so the \(p\)-values should be less than 0.05.

\clearpage
\subsubsection{The Impact of Shadow Dataset and Hyperparameter \(a\)}
\label{app-ablation-alphaDsdw}

\begin{figure}[ht]
    \centering
    \begin{subfigure}{0.49\textwidth}
        \centering
        \includegraphics[width=\textwidth, trim=16cm 0.2cm 5cm 1cm, clip]{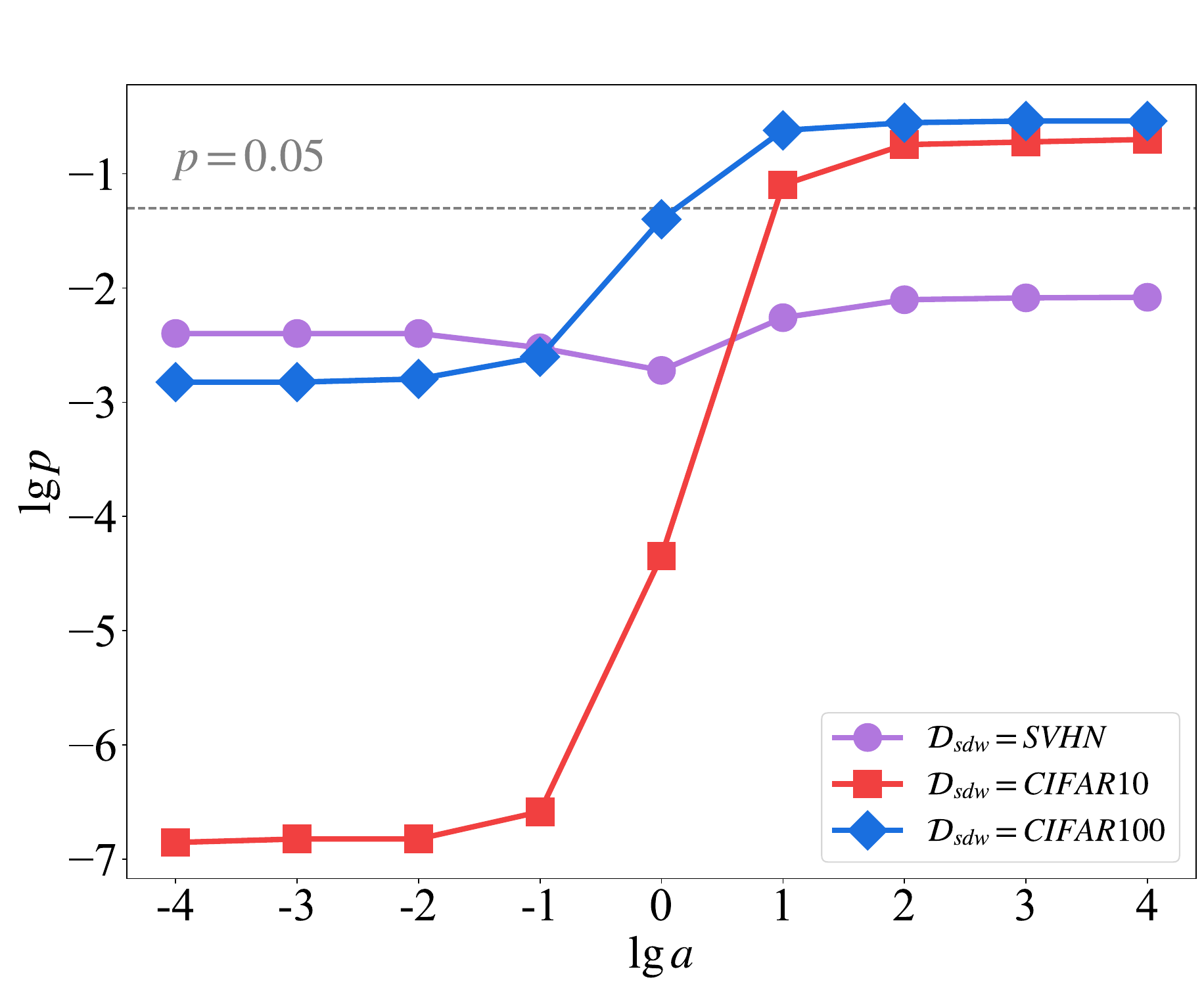}
        \caption{}
        \label{fig:ablation-alphaDsdw-imagenet-simclr}
    \end{subfigure}
    \begin{subfigure}{0.49\textwidth}
        \centering
        \includegraphics[width=\textwidth, trim=16cm 0.2cm 5cm 1cm, clip]{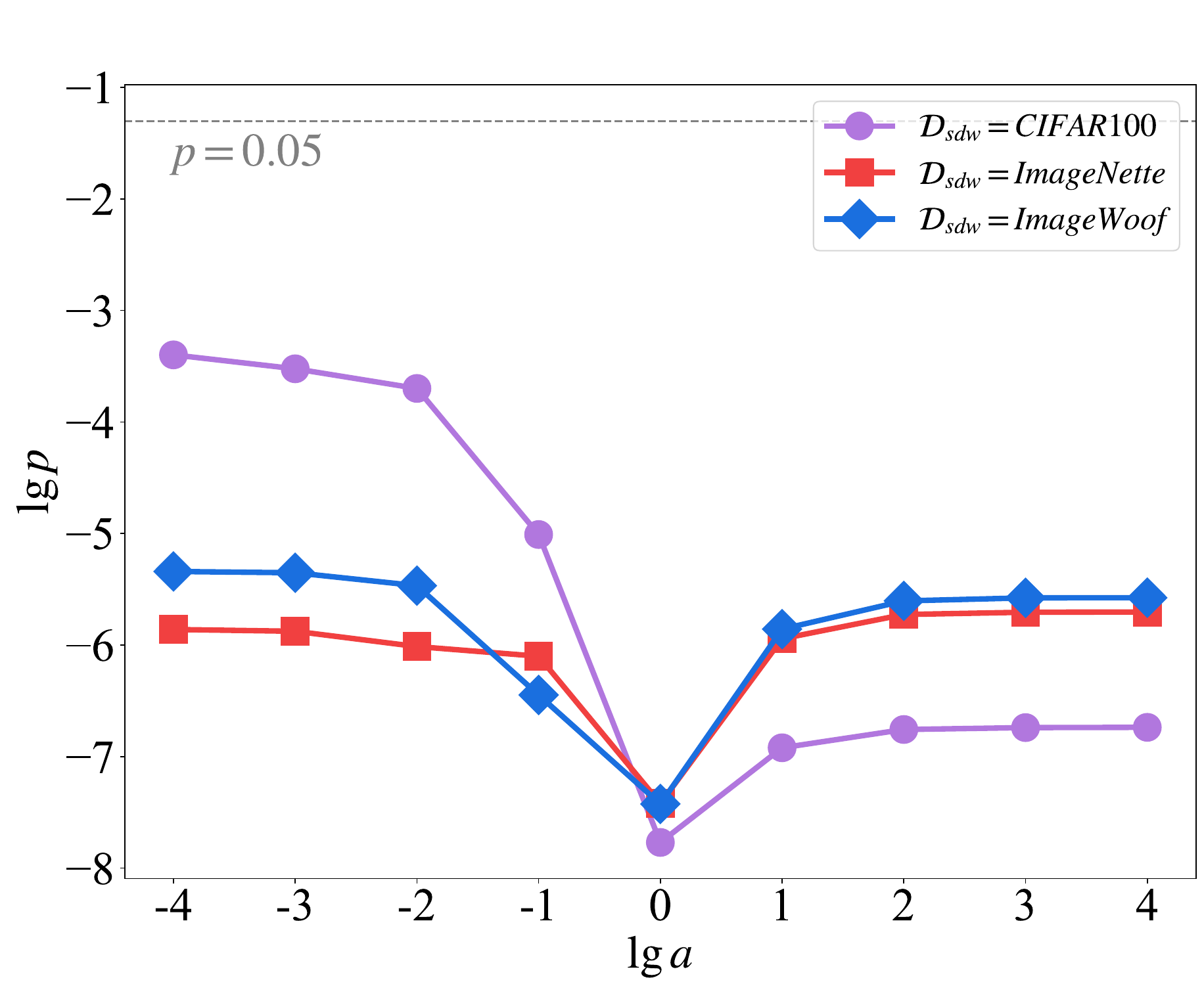}
        \caption{}
        \label{fig:ablation-alphaDsdw-cifar10-simclr}
    \end{subfigure}
    \caption{The impact of shadow dataset and hyperparameter \(a\) on our method. The left figure represent the cases where \(\mathcal{D}_{pub}\) is ImageNet and \(\mathcal{M}_{sus}\) is a pre-trained ResNet50 using SimCLR, and in the right figure,  \(\mathcal{D}_{pub}\) and  \(\mathcal{M}_{sus}\) are CIFAR10 and a pre-trained ResNet18 using SimCLR, respectively.}
    \label{fig:ablation-alphaDsdw}
\end{figure}

We investigate the impact of the shadow dataset \(\mathcal {D}_{sdw}\) and the hyperparameter \(a\) on our method. As shown in Figure \ref{fig:ablation-alphaDsdw}, the setting of \(a\) affects the validation results of our method. This effect is related to the distributions of \(\mathcal {D}_{pub}\) and \(\mathcal {D}_{sdw}\) and is not fixed. This indicates that the defender need to set appropriate \(a\) based on his actual situation.

\subsubsection{The Impact of Early Stopping on Our Method}
\label{app-early}
The early stopping technique can terminate model training prematurely, which may result in less pronounced contrastive relationship gap. To investigate the impact of early stopping on our method, we specifically set the patience of early stopping (the maximum number of epochs allowed to continue training when the K-Nearest Neighbors accuracy on the validation set does not improve significantly over multiple consecutive epochs) to 15 and 30, respectively. We then calculated the $p$-values of the trained models using the same method, as shown in Table \ref{early-table}. Both the datasets of defender and suspect are CIFAR10, meaning $p$-value should be less than 0.05. Self-supervised method is SimCLR. The shadow model is a ResNet18 pre-trained on ImageWoof using SimCLR. The results demonstrate that our method remains effective even under early stopping conditions.

\begin{table}[ht]
\caption{The results (\(p\)-values) of our method on suspicious models that used early stopping. The datasets of defender and suspect are both CIFAR10, making the suspect's behavior illegal, so the \(p\)-values should be less than 0.05.}
\label{early-table}
\centering

\adjustbox{width=\textwidth}{
\begin{tabular}{cccc}

\toprule
 {Model}& {w/o Early Stopping }&{w/ Early Stopping (patience=15)} &{w/ Early Stopping (patience=30)}\\  \hline

 ResNet18& $10^{-12}$ & 0.01 & $10^{-4}$\\
 VGG16 & $10^{-11}$ & $10^{-4}$ & $10^{-5}$\\
 \bottomrule
\end{tabular}}
\end{table}

\clearpage
\subsubsection{The Comparison of Our Method with the Watermark-based Method}
\label{app-compare-bw}
Currently, there is no watermark-based dataset ownership verification method for pre-trained encoders, so we adapt CTRL~\cite{li2023embarrassingly}, the current state-of-the-art backdoor attack for self-supervised encoders, into a watermark-based dataset ownership verification method. Specifically, prior to the release of public dataset, we inject the the CTRL trigger as watermark into a small subset of the data. During the verification phase, we input both watermarked and non-watermarked images into the suspicious encoder. If the representations of the watermarked images are significantly more similar to each other than those of the non-watermarked images, we can conclude that the suspicious encoder was pre-trained using the public dataset. Table \ref{bd-table} indicates that although methods based on watermark can accurately identify cases where public datasets have been stolen, they also wrong the innocent suspect. 

\begin{table}[ht]
\caption{The comparison of our method with watermark-based method (\( \mathcal {D}_{pub} \) is CIFAR10).  `\( \mathcal {D}_{sus} \)' is the dataset used to pre-train \( \mathcal {M}_{sus} \), `Alg' is the dataset ownership verification method. Each value is an average of 3 trials.}
\label{bd-table}
\centering

\begin{tabular}{ccccccc}

\toprule
   \multirow{2}{*}{Alg}& \multirow{2}{*}{Method} &\multirow{2}{*}{Model}&\multicolumn{4}{c}{\( \mathcal {D}_{sus} \)} \\ \cline{4-7} &&& {CIFAR10} &{SVHN} & {ImageNette} & {ImageWoof}  \\ \hline
                          
  \multirow{6}{*}{DOV-CTRL}& \multirow{2}{*}{SimCLR}&VGG16&                               \(10^{-317}\)&                             \(10^{-4}\)&                               \(10^{-6}\)&                          \(10^{-49}\)\\
 & & ResNet18&   \(10^{-287}\)&   \(10^{-6}\)&   0.04& 0.19\\ \cline{2-3}
  & \multirow{2}{*}{BYOL}&VGG16&   0&   0.02&   0.57&  0.47\\
 & & ResNet18&   0&   0.29&   0.52&  0.04\\ \cline{2-3}
  & \multirow{2}{*}{SimSiam}&VGG16
&   \(10^{-19}\)&   0.17&   \(10^{-8}\)&  \(10^{-4}\)\\
 & & ResNet18&   \(10^{-290}\)&   0.25&   0.18& 0.31\\ \hline       \multirow{6}{*}{Ours}& \multirow{2}{*}{SimCLR}
&VGG16
&                                     \(10^{-10}\)&                            0.45&                              0.67&                          0.99\\
 & 

& ResNet18&   \(10^{-7}\)&   0.41&  0.19&  0.16\\  \cline{2-3}

  & \multirow{2}{*}{BYOL}
&VGG16
&   \(10^{-13}\)&   0.84&   0.85&  0.90\\
 & 

& ResNet18&   \(10^{-9}\)&   0.61&  0.59& 0.59\\  \cline{2-3}
  & \multirow{2}{*}{SimSiam}
&VGG16
&   \(10^{-4}\)&   0.99&   0.98&  0.98\\
 & & ResNet18&  \(10^{-4}\)&   0.88&   0.36&  0.37\\
 \bottomrule
\end{tabular}
\end{table}

\subsubsection{The Performance of Our Method on MAE}
\label{app-mae}
We also conduct experiments using encoders pre-trained with methods other than contrastive learning. We select Masked Autoencoder (MAE)~\cite{he2022masked} for experimentation, which is a representative method of Masked Image Modeling (MIM). Specifically, we use pre-trained models on ImageNet from the official MAE repository\footnote{\href{https://github.com/facebookresearch/mae}{https://github.com/facebookresearch/mae}}, with encoder architectures ViT-B/16 and ViT-L/16. Additionally, to better adapt the encoders pre-trained with MIM, we incorporate random masking into our multi-scale augmentation. According to the experimental results presented in Table \ref{mae-table}, our method still didn't perform well despite targeted improvement. We will address the DOV issue of MIM pre-trained models in our future work.

\begin{table}[ht]
\caption{The results (\(p\)-values) of our method on MAE (\( \mathcal {D}_{pub} \) is ImageNet).  `\( \mathcal {D}_{sus} \)' is the dataset used to pre-train \( \mathcal {M}_{sus} \). Each value is an average of 3 trials. Note that in this scenario, \(\mathcal {D}_{sus}\) and \(\mathcal {D}_{pub}\) are both ImageNet and \(\mathcal {D}_{sus}\) includes \(\mathcal {D}_{pub}\), making the suspect's behavior illegal, so the \(p\)-values should be less than 0.05.}
\label{mae-table}
\centering

\begin{tabular}{ccc}

\toprule
 {Method}& Model&{Ours + Random Masking}\\  \hline

 \multirow{2}{*}{MAE}& ViT-B/16&0.75\\
  & ViT-L/16& 0.44\\
 \bottomrule
\end{tabular}
\end{table}

\subsubsection{Visualization Results}
\label{app-visual}

\begin{figure}[ht]
    \centering
    \begin{subfigure}{0.49\textwidth}
        \centering
        \includegraphics[width=\textwidth, trim=10cm 6cm 4cm 2cm, clip]{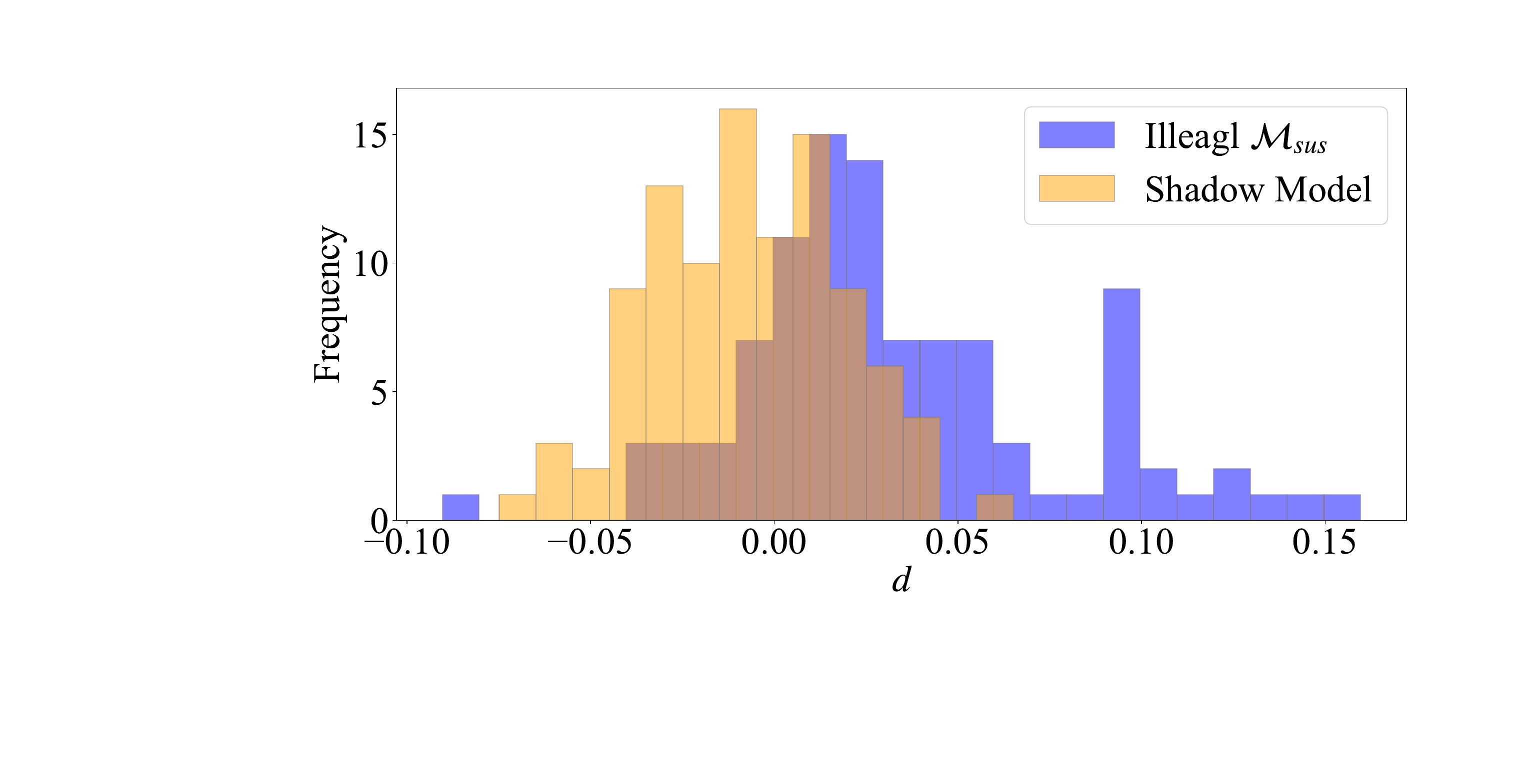}
        \caption{}
        \label{fig:nette-simclr}
    \end{subfigure}
    \hfill
    \begin{subfigure}{0.49\textwidth}
        \centering
        \includegraphics[width=\textwidth, trim=10cm 6cm 4cm 2cm, clip]{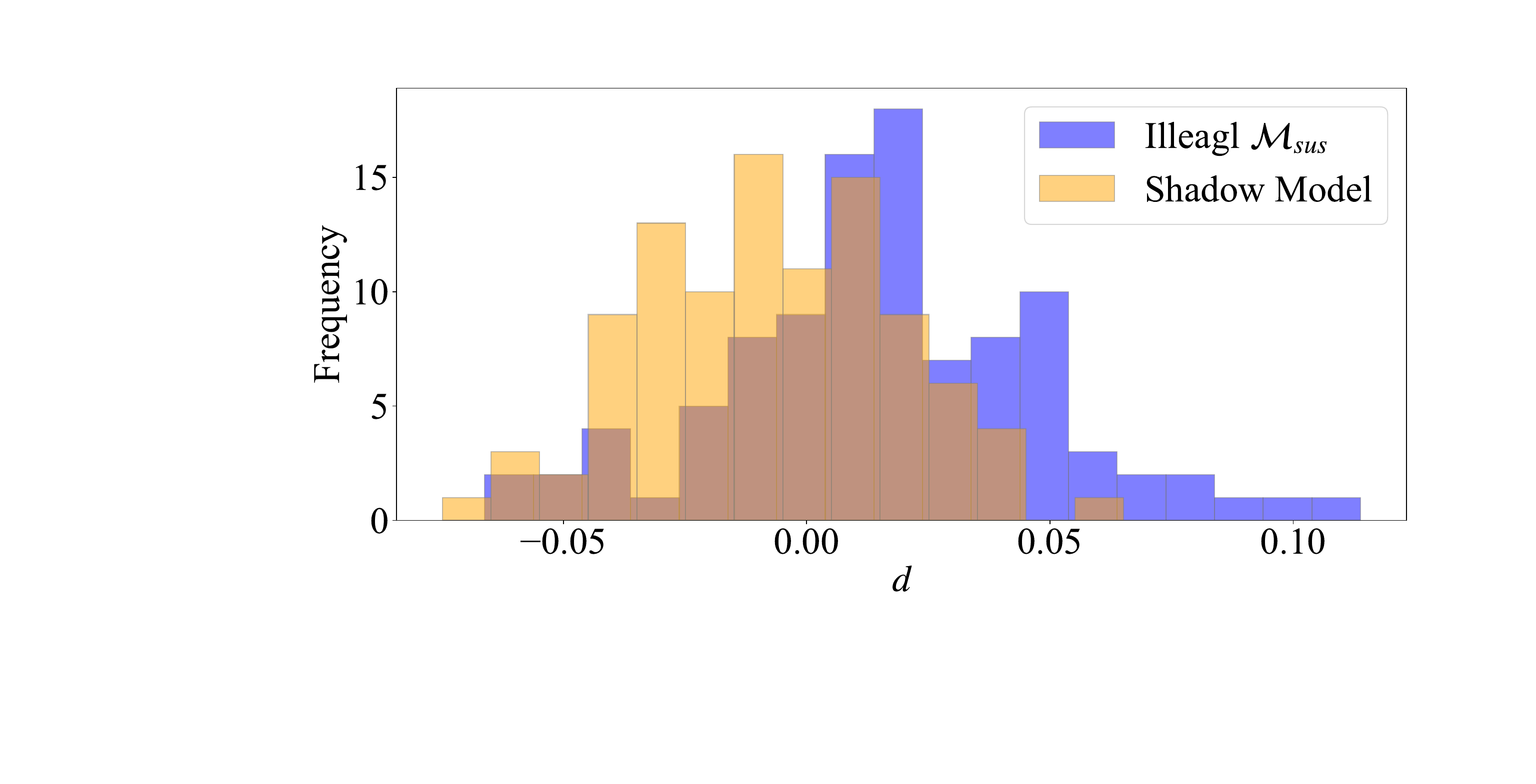}
        \caption{}
        \label{fig:nette-simsiam}
    \end{subfigure}
    \hfill
    \begin{subfigure}{0.49\textwidth}
        \centering
        \includegraphics[width=\textwidth, trim=10cm 6cm 4cm 2cm, clip]{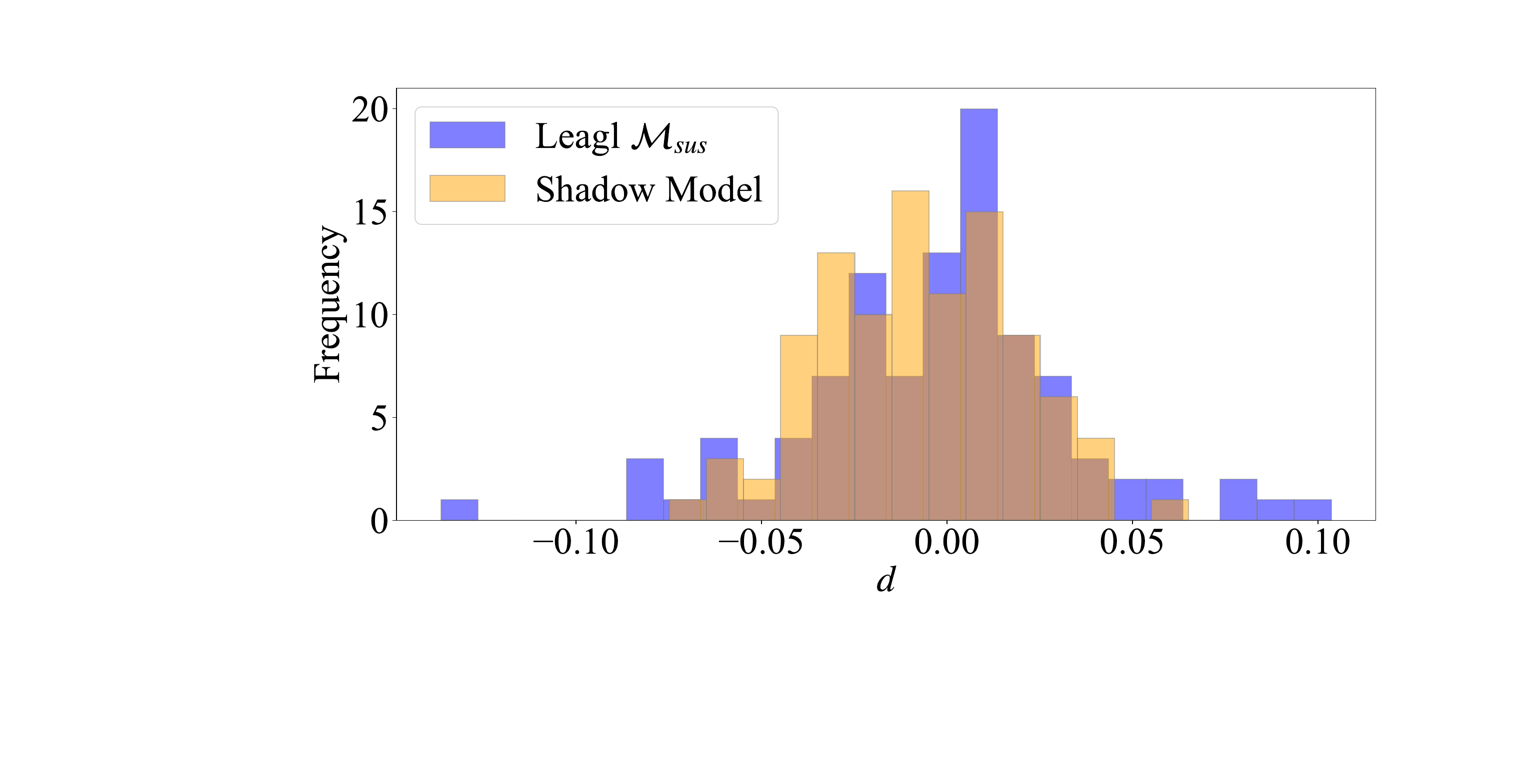}
        \caption{}
        \label{fig:woof-simclr}
    \end{subfigure}
    \hfill
    \begin{subfigure}{0.49\textwidth}
        \centering
        \includegraphics[width=\textwidth, trim=10cm 6cm 4cm 2cm, clip]{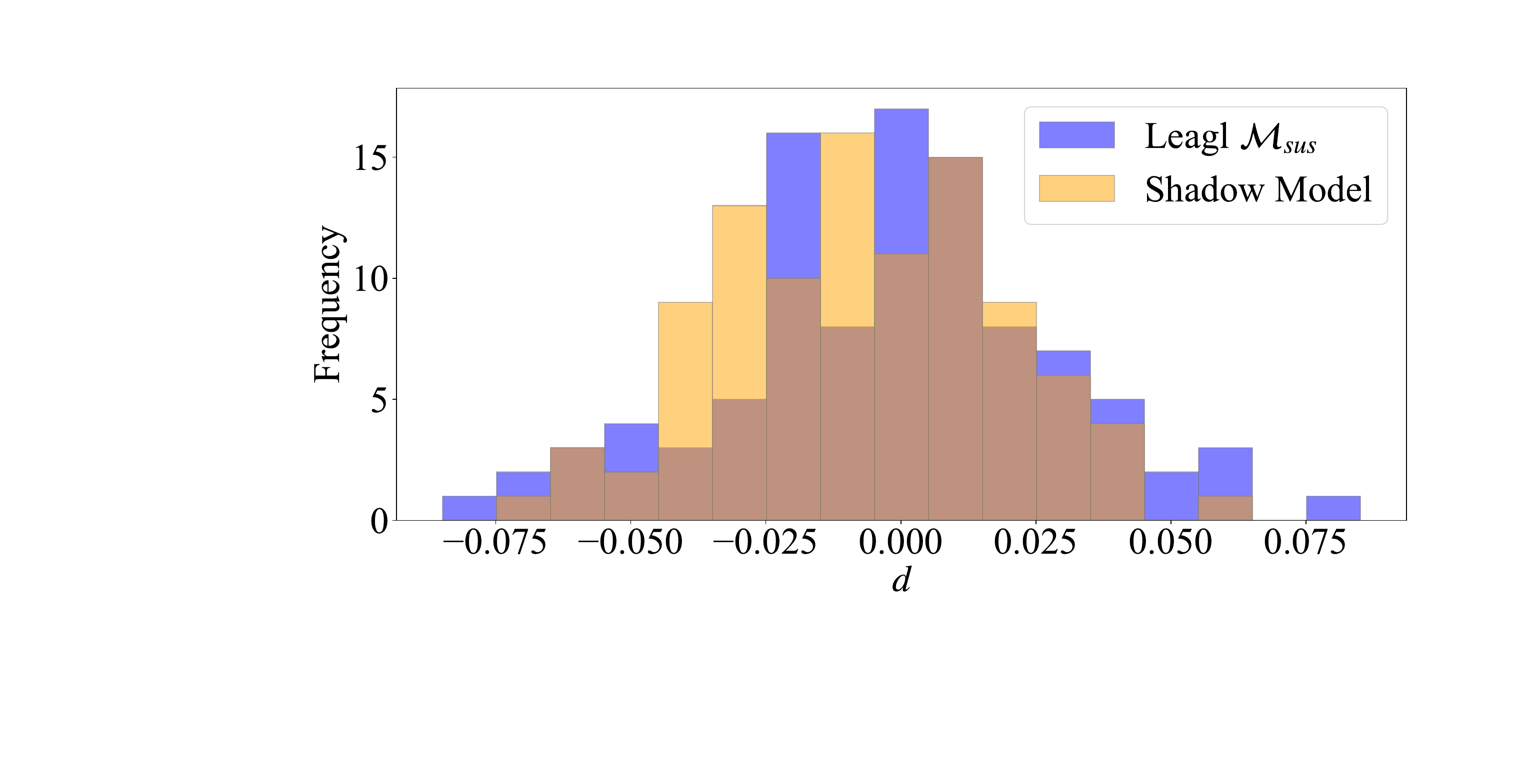}
        \caption{}
        \label{fig:woof-simsiam}
    \end{subfigure}
    \caption{The contrastive relationship gap \( d \) of the shadow model and suspicious models trained on different datasets. Each subplot corresponds to a different suspicious model. Figure \ref{fig:nette-simclr}: suspicious model is a ResNet18 trained on ImageNette using SimCLR, Figure \ref{fig:nette-simsiam}: suspicious model is a ResNet18 trained on ImageNette using SimSiam, Figure \ref{fig:woof-simclr}: suspicious model is a ResNet18 trained on ImageWoof using SimCLR, and Figure \ref{fig:woof-simsiam}: suspicious model is a ResNet18 trained on ImageWoof using SimSiam.}
    \label{fig:visual}
\end{figure}

We present the visualization results of our method on ImageNette. Specifically, \( \mathcal{D}_{pub} \) is set as ImageNette, and the shadow model is a ResNet18 trained on SVHN using SimCLR. We calculated the contrastive relationship gap \( d \) of the shadow model and suspicious models trained on different datasets and visualized the comparison. 
When the suspicious model is pre-trained on \( \mathcal{D}_{pub} \), it is considered illegal, and the contrastive relationship gap \( d \) should be significantly higher than that of the shadow model. Conversely, if the suspicious model is legitimate, the two contrastive relationship gaps should be similar. 
As shown in Figure \ref{fig:visual}, when the suspicious model is illegal, its contrastive relationship gap is significantly higher than that of the shadow model. When the suspicious model is legitimate, the two contrastive relationship gaps are close. This observation aligns with our previous findings.

\end{document}